\pdfoutput=1

\documentclass[11pt]{article}

\usepackage[final]{acl}

\usepackage{times}
\usepackage{latexsym}

\usepackage[T1]{fontenc}

\usepackage[utf8]{inputenc}

\usepackage{microtype}

\usepackage{inconsolata}

\usepackage{amsmath}
\usepackage{amsfonts}
\usepackage{graphicx}
\usepackage{multirow}
\usepackage{booktabs}
\usepackage{subfigure}
\usepackage{longtable}
\usepackage{tabularx}

\title{Training Language Model to Critique for Better Refinement}

\author{
Tianshu Yu$^{1,2*}$, Chao Xiang$^{1}$\thanks{\ \ Equal contribution.}, Mingchuan Yang$^{1}$, Pei Ke$^{3}$, Bosi Wen$^{2\dagger}$, Cunxiang Wang$^{4,5}$, \\  
\textbf{Jiale Cheng$^{2}$\thanks{\ This work was conducted when Bosi Wen and Jiale Cheng were interning at Zhipu AI.}, Li Zhang$^{1}$, Xinyu Mu$^{1}$, Chuxiong Sun$^{1}$, Minlie Huang$^{2}$\thanks{\ Corresponding author.}}  \\
  $^{1}$China Telecom Research Institute \\ 
  $^{2}$The Conversational Artificial Intelligence (CoAI) Group, Tsinghua University\\
  $^{3}$University of Electronic Science and Technology of China \\
  $^{4}$The Knowledge Engineering Group (KEG), Tsinghua University \\
  $^{5}$Zhipu AI \\
  \texttt{dailyyulun@gmail.com}   \ \ \ \ \ \ \texttt{aihuang@tsinghua.edu.cn}
}

\begin{document}
\maketitle
\begin{abstract}
Large language models (LLMs) have demonstrated remarkable evaluation and critique capabilities, providing insightful feedback and identifying flaws in various tasks. However, limited research has explored which types of critiques are most effective for improving model responses or how to generate such critiques.
To address this gap, we introduce \textbf{R}efinement-oriented \textbf{C}ritique \textbf{O}ptimization (RCO), a novel framework designed to train critic models using refinement signals. 
RCO uses a feedback loop where critiques, generated by the critic model, guide the actor model in refining its responses. The critique utility (CU) quantifies the effectiveness of these refinements, serving as the reward signal for training the critic model. By focusing on critiques that lead to better refinements, RCO eliminates the need for direct critique preference assessment, ensuring that critiques driving meaningful improvements are rewarded.
We evaluate RCO across five tasks, i.e., dialog generation, summarization, question answering, mathematical reasoning, and code generation, and show that it significantly outperforms traditional methods and open-source models in terms of critique quality and refinement outcomes. Our contributions include the introduction of RCO, a novel supervision scheme based on refined response preferences, and comprehensive experimental results that highlight the method's effectiveness in enhancing LLM critique-refinement loops.\footnote{We release the code and data at https://github.com/publicstaticvo/critique.}

\end{abstract}

\section{Introduction}
\label{sec:intro}

The critique ability, defined as the capacity to identify and refine flaws in responses, is crucial for improving the reliability of automatic evaluation and enabling the self-improvement of large language models (LLMs)~\cite{saunders2022self,scheurer2023training}. 
Enhancing this ability is challenging, as it requires LLMs to not only detect errors but also engage in a deeper analysis of user queries and responses, extending beyond superficial criticism~\cite{zheng2023codegeex,kim2024prometheus}. 
Recent studies have sought to address this challenge of LLMs via human-curated critique datasets and alignment strategies like supervised fine-tuning (SFT)~\cite{cui2023ultrafeedback,ke2024critiquellm} and reinforcement learning~\cite{akyurek2023rl4f,mcaleese2024llm}, yielding promising results in identifying and rectifying shortcomings across various tasks.

\begin{figure}[t]
    \centering
    \includegraphics[width=1.0\linewidth]{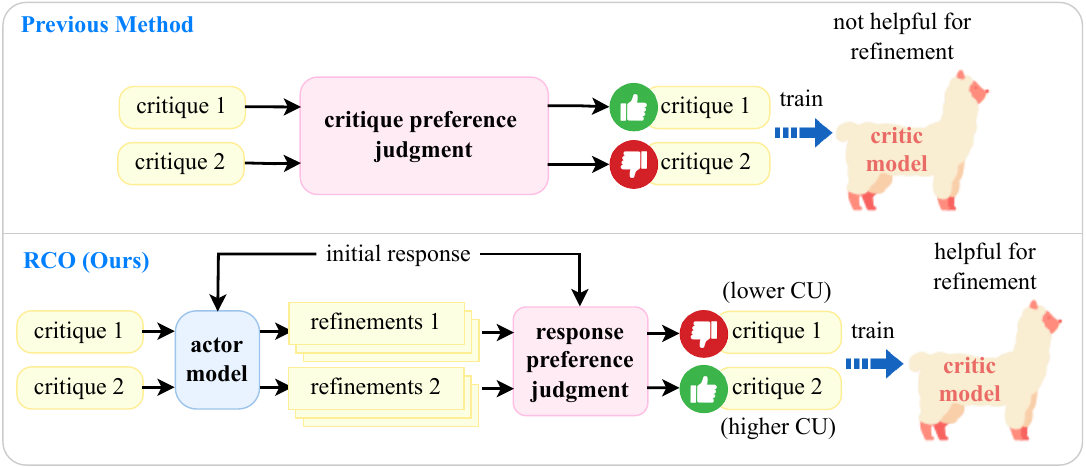}
    \caption{A comparison between previous methods, which train critic models via direct critique preference judgment and are not helpful for refinement, and our method, which train critic models by converting preferences of refined responses into critique reward values and is helpful for refinement.}
    \label{fig:pre}
\end{figure}

However, current methods primarily train models to generate critiques for evaluation purposes, rather than linking critique to refinement~\cite{li2024llms}. 
This disjointed approach fails to connect error identification with output improvement, limiting the practical value of critiques.
While the critique-refinement paradigm is explored in LLM alignment~\cite{madaan2023self,wadhwa2024learning} and reasoning~\cite{chen2024magicore,wu2024comparative}, 
existing works lack systematic investigation on how critique enhancements can lead to meaningful refinements in responses.
Without this link, assessing critique quality remains challenging
~\cite{sun2024critique}. 
Addressing this gap is essential for advancing both LLM evaluation and self-improvement capabilities.

To address these challenges, we introduce Refinement-oriented Critique Optimization (RCO), a novel training paradigm for critic models that uses critique utility (CU) as a reward signal, which is calculated by comparing the refined responses to the initial ones.
This approach encourages the generated critiques
to drive substantial improvements in the output, creating a more effective critique-refinement loop. 
Specifically, RCO feeds critiques and initial responses into the actor model and prompts it to generate multiple refined responses. 
CU is then quantified as the proportion of refinements preferred over the initial response, which serves as the reward signal for training the critic model. 
By focusing on the refinement outcomes, RCO eliminates the need for direct critique preference assessment and ensures that critiques leading to better refinements are rewarded.
We evaluate RCO across five tasks, i.e., dialog generation, summarization, question answering, mathematical reasoning, and code generation, using baseline models, models trained with critique preferences, and advanced open-source models. Experimental results demonstrate that RCO significantly enhances critic model performance, outperforming existing methods across multiple benchmarks.

Our contributions are threefold:
(1) We propose RCO, a method that prioritizes critiques that facilitate effective refinement of actor model responses, addressing limitations of previous approaches.
(2) We introduce a novel supervision scheme based on refined response preferences, eliminating the need for directly assessing the quality of critiques while rewarding critiques that lead to meaningful improvements.
(3) We rigorously evaluate RCO across diverse tasks, showing substantial improvements in critique quality and refinement capabilities, and provide an in-depth analysis of the method's impact.

\begin{figure*}[t]
    \centering
    \includegraphics[width=1.0\linewidth]{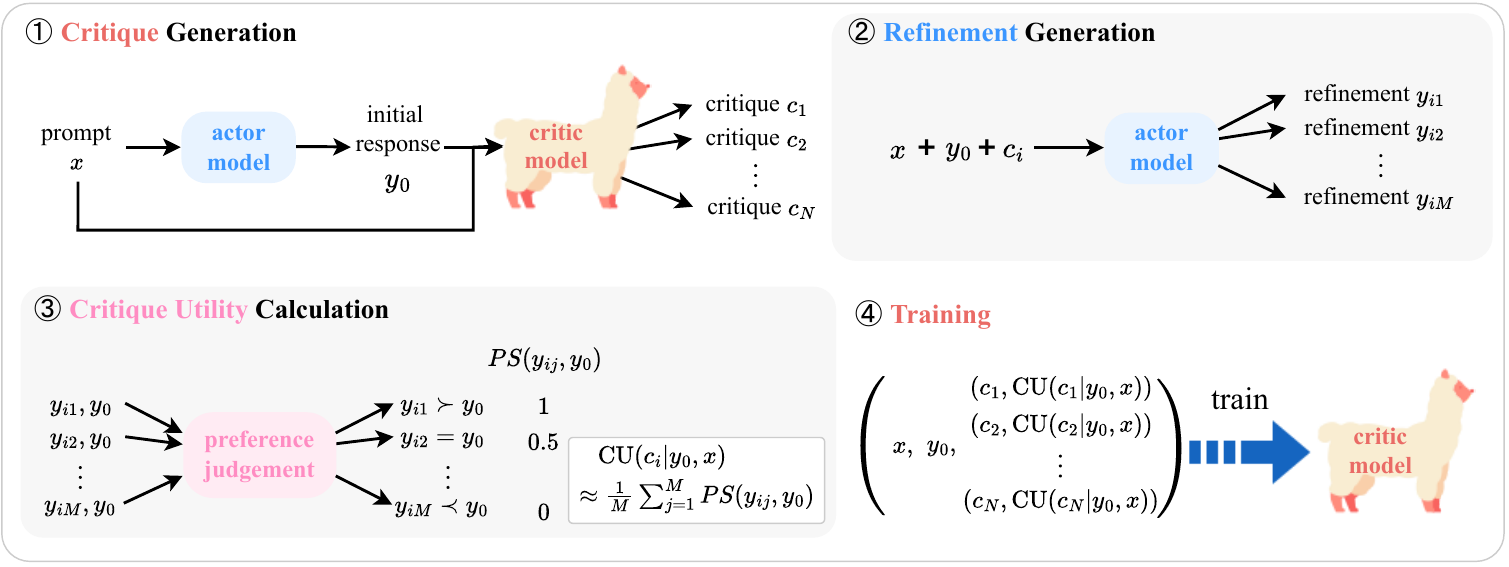}
    \caption{The illustration of our method RCO, describing our data collection and training process. We first generate critiques with the critic model, then sample multiple refined responses with the actor model. Subsequently, we calculate the critique utility of each critique with preference judgment on refined and initial responses, which serves as reward signal of critic model training.}
    \label{fig:main}
\end{figure*}

\section{Related Work}

\paragraph{Critique Ability of LLMs} The rapid advancement of large language models (LLMs) has highlighted the need to enhance their critique capabilities. LLMs like GPT-4~\cite{achiam2023gpt} have proven effective as evaluators~\cite{zheng2023judging,li2024generation,cao2024compassjudger}, but their API-based access limits widespread application. To address cost and stability issues, researchers have fine-tuned open-source models using critique data from these models. Despite these efforts, challenges in critique task complexity remain. Recent work includes \citet{murugadoss2024evaluating}, which uses prompt engineering to create critique metrics, and \citet{verga2024replacing}, which proposes using multiple LLMs to mitigate bias. \citet{ke2024critiquellm} generates a golden critique dataset from model and reference response pairs, while \citet{lan2024training} employs a multi-agent framework for preference-based critique data collection. In contrast, our method improves critique quality by automatically obtaining preferences of refined responses to calculate critique utility as reward signals to train critic models.

\paragraph{Preference-Based Reinforcement Learning} Reinforcement Learning from Human Feedback (RLHF)~\cite{ziegler2019fine} is commonly used to guide LLMs towards human-preferred responses. \citet{scheurer2023training} uses RLHF to train reward models based on human-annotated pairwise comparisons. Recent methods, such as CriticGPT~\cite{mcaleese2024llm}, apply RLHF to enhance critique abilities by training critic models on human-identified errors in code generation. Similarly, \citet{wang2024self} collects preference critique pairs by comparing LLM-generated responses with human-annotated scores. These approaches are limited by the high cost and uncertain quality of human annotations~\cite{sun2024critique}. In contrast, our approach reduces reliance on human annotations, providing a clear standard for good critiques that effectively guide actor model refinements.

\section{Methodology}
\label{sec:method}

The overview of RCO is illustrated in Figure \ref{fig:main}. 
Our approach begins with a dataset \(\mathcal{D}\), where each sample \(\mathbf{X} = (x, y_0)\) consists of a prompt \(x\) and its corresponding initial response \(y_0\), with \(y_0\) being generated by an actor model \(\pi(y_0 | x)\). Following this, a base critic model \(p(c | y_0, x)\) is employed to generate \(N\) distinct critiques, denoted as \(c_1, c_2, \dots, c_N\), for the initial response \(y_0\). Each critique \(c_i\) is then used as input to the actor model \(\pi\), which produces \(M\) distinct refined responses, \(y_{i1}, y_{i2}, \dots, y_{iM}\). The distribution used to sample these refined responses is represented as \(\pi_{c_i}(y | c_i, y_0, x)\). 


\paragraph{Critique Utility.}
As highlighted in Section \ref{sec:intro}, the critique process is increasingly integrated into critique-refinement workflows, requiring supervision from a refinement perspective. Inspired this, we propose to use critique utility (CU) as the reward signal of RCO.
The central idea of CU is that a high-quality critique should naturally lead to a better refinement, serving as a catalyst for self-improvement~\cite{sun2024critique}. 
Specifically, CU for a critique $c_i$ is defined as the probability where a response $y$, sampled from $\pi_{c_i}(y | c_i, y_0, x)$, is preferred over the initial response $y_0$:
\begin{equation}
    \text{CU}(c_i | y_0, x) = P(y \succ y_0|y \sim \pi_{c_i})
\end{equation}
To estimate $\text{CU}(c_i | y_0, x)$, we approximate it using the following sampling-based approach, by defining a preference score (PS) for the refined response $y_{ij}$ and the initial response $y_0$:
\begin{equation}
\label{eq:r}
    \text{CU}(c_i | y_0, x) \approx \frac{1}{M} \sum_{j=1}^M \text{PS}(y_{ij}, y_0)
\end{equation}
In this equation, the preference score \( \text{PS}(y_{ij}, y_0) \) is determined by a judge model based on the following criteria: \( \text{PS}(y_{ij}, y_0) = 1 \) if the refined response \( y_{ij} \) is preferred over the initial response \( y_0 \), \( \text{PS}(y_{ij}, y_0) = 0.5 \) if both responses are considered equally good, and \( \text{PS}(y_{ij}, y_0) = 0 \) if \( y_0 \) is preferred over \( y_{ij} \). This process is repeated for each critique \( c_i \).

\paragraph{Training Objective Derivation.}
The goal of RCO is to reward the generated critique by critique utility. We derive our training objective starting from the following optimization problem:

\begin{footnotesize}
\begin{equation}
\label{eq:ppo}
    \max_{\theta}\mathbb{E}_{c\sim p_\theta}\left[R(c|y_0,x)\right]-\beta\mathbb{D}_{\text{KL}}\left[p_\theta(c|y_0,x)\|p(c|y_0,x)\right]
\end{equation}
\end{footnotesize}
where $R(c|y_0,x)$ is the reward function of the critique $c$ given the prompt $x$ and initial response $y_0$. Since we use critique utility as the reward signal, we substitute $R(c|y_0,x)$ with $\text{CU}(c|y_0,x)$ for the rest of the derivation. 
According to \citet{rafailov2023direct}, the optimal solution of Eq.\ref{eq:ppo} takes the following form:
\begin{equation}
    p^*(c|y_0,x)=\frac{p(c|y_0,x)\exp{\left(\frac{1}{\beta} \text{CU}(c | y_0, x)\right)}}{Z_\beta(y_0, x)}
\end{equation}
where $Z_\beta(y_0, x)$ is the regularization term. With some algebra, we obtain:
\begin{equation}
    \frac{1}{\beta}\text{CU}(c|y_0,x)=\log\frac{p^*(c|y_0,x)}{p(c|y_0,x)}+\log Z_\beta(y_0, x)
\end{equation}
Note that the regularization term $Z_\beta(y_0, x)$ can be approximated via sampling in our method:
\begin{equation}
\label{eq:z}
\begin{aligned}
    Z_\beta(y_0, x) &= \mathbb{E}_{c\sim p(c | y_0, x)} \exp\left(\frac{1}{\beta} \text{CU}(c | y_0, x)\right)\\
    &\approx \frac{1}{N} \sum_{i=1}^N \exp\left(\frac{1}{\beta} \text{CU}(c_i | y_0, x)\right)
\end{aligned}
\end{equation}
Finally, inspired by Direct Reward Optimization~\cite{richemond2024offline}, we utilize a minimum square error objective as the training objective of our method, substituting $p^*(c_i | y_0, x)$ with $p_{\theta}(c_i | y_0, x)$, the model to be optimized:
\begin{equation}
\label{eq:loss}
\begin{aligned}
    \mathcal{L}_{\text{RCO}} &= \mathbb{E}_{(x,y_0)\in\mathcal{D}}\Bigg[\frac{1}{2N} \sum_{i=1}^N \bigg(\log \frac{p_{\theta}(c_i | y_0, x)}{p(c_i | y_0, x)}  \\ & + \log Z_\beta(y_0, x) - \frac{1}{\beta}\text{CU}(c_i | y_0, x) \bigg)^2\Bigg]
\end{aligned}
\end{equation}
Compared to traditional preference-based learning methods such as DPO~\cite{rafailov2023direct}, our training objective leverages scalar reward values more effectively, enabling the critic model to learn a more nuanced reward representation. This advantage allows our method to better capture complex reward structures, thereby improving overall model performance.

\subsection{Training Data Collection}
\label{sec:data}

\paragraph{Dataset Overview.} Our method begins with the collection of the prompt dataset $\mathcal{D}$ consisting of five tasks: dialog generation, summarization, question answering, mathematical reasoning, and code generation. These tasks are sourced from 14 datasets, as outlined in Table \ref{tab:dataset}, with a total of 10,000 unique prompts collected for the experiment.

\paragraph{Collection of Initial Responses.} Four actor models, i.e., \textit{LLaMA-2-7B-Chat}, \textit{LLaMA-2-13B-Chat}, \textit{LLaMA-2-70B-Chat}, and \textit{LLaMA-3-8B-Instruct}~\cite{touvron2023llama,dubey2024llama}, are used to generate responses for the 10,000 prompts, yielding 40,000 unique responses. Among these, 8,000 responses are selected per model, ensuring 1,600 responses per task and 2,000 per actor model.

\paragraph{Critique Generation.} To generate critiques, we employ five base critic models: \textit{LLaMA-2-7B-Chat}, \textit{LLaMA-2-13B-Chat}, \textit{LLaMA-3-8B-Instruct}, \textit{Auto-J-13B}~\cite{li2023generative}, and \textit{UltraCM-13B}~\cite{cui2023ultrafeedback}. Each model generates $N=4$ critiques for every initial response $y_0$ in the training set, ensuring a more accurate approximation of the regularization term $Z_\beta(y_0, x)$ and improving the training procedure.

\paragraph{Refinement Generation Based on Critiques.} In the refinement phase, the actor model that generates the initial response $y_0$ refines its output based on the critique it receives. For each critique $c_i$, the actor model generates $M=5$ distinct refined responses, which enables a better approximation of critique utility.

\paragraph{Critique Utility Calculation.} As shown in Figure \ref{fig:main}, critique utility $\text{CU}(c_i|y_0,x)$ is calculated as the reward signal for each critique, following the method in Section \ref{sec:method}. We use the \textit{Qwen-2.5-72B-Instruct}~\cite{qwen2.5} model to evaluate the preference of each refined response $y_{ij}$ relative to its initial response $y_0$. To avoid positional bias, we alternate the positions of the refined and initial responses and perform response preference judgment again \cite{wang2024faireval}. The critique utility for each critique $c_i$ is computed as the average of 10 individual preferences ($2M=10$). This process ensures robust evaluation while mitigating potential bias, with data collection prompts detailed in Appendix \ref{app:dataprompt}.

\section{Experiment Settings}

In this section, we describe the benchmarks, evaluation metrics, baselines, and experimental results that assess the performance of RCO.

\subsection{Benchmarks and Evaluation Metrics}
\label{sec:metric}

\paragraph{Critique Utility and Refinement Quality Evaluation.} 
We create a test dataset of 2,500 prompts, sourced from 7 datasets distinct from those in the training set, as detailed in Table \ref{tab:testset}. 
For initial responses, we use five actor models: the four models from the training dataset and \textit{LLaMA-3-70B-Instruct}, a larger model not used for the training dataset.
After generating critiques with the critic model, we evaluate its quality by greedy sampling a single refined response from the actor model of the initial response for each critique.
We use two metrics: (1) average \textbf{CU}, calculated by prompting GPT-4 to compare the preference of refined responses to initial responses, and (2) \textbf{response quality score (RQS)}, where GPT-4 rates the responses on a 1-10 scale, with higher scores indicating better quality. We provide the prompts used for evaluation in Appendix \ref{app:evalprompt}.

\paragraph{Refinement Accuracy Evaluation.} 
To quantitatively assess the impact of critique-based refinement, we evaluate refined response \textbf{accuracy} across 10 established benchmarks spanning three critical reasoning domains: (1) \textbf{Question Answering}: BBH~\cite{suzgun2022challenging}, GPQA-Diamond~\cite{rein2024gpqa}, MMLU~\cite{hendrycks2020measuring}, MMLU-Pro~\cite{wang2024mmlu}, MMLU-Redux~\cite{gema2024we}, and TruthfulQA~\cite{lin2021truthfulqa}. (2) \textbf{Mathematical Reasoning}: MATH~\cite{hendrycks2021measuring} and GSM8K~\cite{cobbe2021training}. (3) \textbf{Code Generation}: MBPP~\cite{austin2021program} and MBPP-Plus\cite{cassano2023multipl}.
Our evaluation methodology follows a rigorous three-stage pipeline: (1) initial response generation using the same actor models as in CU/RQS evaluation, (2) comprehensive critique generation for all initial responses, and (3) refined response generation and evaluation. For QA and math tasks, we employ \textit{Qwen2.5-72B-Instruct} for answer consistency checking against gold standards. Code generation tasks are evaluated using the \textit{bigcode-project-harness}~\cite{bigcode-evaluation-harness} testing framework, which provides deterministic verification through test case execution.
This multi-domain evaluation strategy ensures comprehensive assessment of our method's generalization capability across different types of reasoning tasks while maintaining objective, verifiable measurement standards.

\paragraph{Pairwise Critique Evaluation.} 
We further investigate the model's capability as a discriminative judge through rigorous testing on RewardBench~\cite{lambert2024rewardbench}, a standardized benchmark for reward model evaluation. Our evaluation protocol strictly adheres to the benchmark's original design while adapting the judgment generation process as specified in Tables \ref{tab:prompt1}-\ref{tab:prompt5}. 
This evaluation provides crucial insights into the model's ability to function as a general-purpose preference judge model beyond the critique refinement paradigm.

\paragraph{Human Evaluation.} 
The second benchmark combines CriticEval~\cite{lan2024criticbench} and CriticBench~\cite{lin2024criticbench} datasets to evaluate the critique ability of models across tasks. 
Human evaluation assesses critique and refinement quality: (1) \textbf{Human preferences of critiques} compare critiques generated by our method and baselines, and (2) \textbf{Human preferences of refinements} compare refinements generated from both critiques by \textit{LLaMA-2-7B-Chat}. 
We sample 200 responses, ensuring 40 responses per task and disjoint prompt sources from those used in training. Details of the human evaluation process are provided in Appendix \ref{app:human}.

\subsection{Baselines}

We evaluate our method against 5 baseline types: 
(1) \textbf{Base critic models:} The five base models described in Section \ref{sec:data}.
(2) \textbf{Self-refinement:} Following prior works~\cite{akyurek2023rl4f}, we use the actor model to directly refine its own responses in the test set and evaluate the average preference score (PS) and RQS of the refined responses.
(3) \textbf{Open-source LLMs:} We compare against models such as \textit{LLaMA-2-70B-Instruct} and \textit{LLaMA-3-70B-Instruct}.
(4) \textbf{Aligner:} \textit{Aligner}~\cite{ji2024aligner} is a model-agnostic module that refines responses; we use \textit{Aligner-7B-V1.0} and evaluate the average PS and RQS of its refined responses.
(5) \textbf{Direct Preferences of Critique Optimization (DPCO):} We use \textit{Qwen-2.5-72B-Instruct} to assess the preferences of critique pairs in the training dataset, following the process in Figure \ref{fig:pre}. For the $N=4$ setting, we label the preferences of critique pairs $(c_1, c_2)$ and $(c_3, c_4)$ and train critic models using the DPO algorithm.

\begin{table*}[ht]
\resizebox{1.0\textwidth}{!}{
\begin{tabular}{cccc|cc|cc|cc|cc|cc}
\toprule
\multirow{2}{*}{\begin{tabular}[c]{@{}c@{}}Base \\ Model\end{tabular}} & \multirow{2}{*}{Method} & \multicolumn{2}{c}{Dialog} & \multicolumn{2}{c}{Summ.} & \multicolumn{2}{c}{QA.} & \multicolumn{2}{c}{Math} & \multicolumn{2}{c}{Code} & \multicolumn{2}{c}{Overall} \\ \cmidrule{3-14} 
& & CU & RQS & CU & RQS & CU & RQS & CU & RQS & CU & RQS & CU & RQS \\ \midrule
\multicolumn{2}{c}{Initial Answer} & -- & 6.23 & -- & 6.52 & -- & 5.33 & -- & 3.73 & -- & 4.42 & 46.9 & 5.25\\ \midrule
\multicolumn{14}{c}{BASELINES}  \\ \midrule
\multicolumn{2}{c}{LLaMA-2-70B-Chat} &  82.7 &7.09 & 68.3 &7.44 & 88.8 &6.75 & 62.7 &4.19 & 59.7 &5.59 & 72.4 &6.21 \\
\multicolumn{2}{c}{LLaMA-3-70B-Instruct} & 82.6 &7.08 & 87.8 &7.68 & 86.3 &6.47 & 76.2 &4.75 & 78.1 &6.24 & 82.2 &6.44 \\
\multicolumn{2}{c}{Self-refinement} & 75.2 &6.96 & 77.6 &7.46 & 79.9 &6.08 & 64.5 &4.32 & 65.8 &5.06 & 72.6 & 5.97 \\
\multicolumn{2}{c}{Aligner} & 47.3 & 6.30 & 50.2 & 6.64 & 47.6 & 5.79 & 51.4 & 3.78 & 38.0 & 4.49 & 46.9 & 5.40 \\ \midrule
\multicolumn{14}{c}{OUR METHOD}\\ \midrule
\multirow{3}{*}{\begin{tabular}[c]{@{}c@{}}LLaMA-2-\\7B-Chat\end{tabular}}     & Base model & 83.5 &7.16 & 63.4 &7.43 & 87.1 &6.48 & 60.1 &4.03 & 59.7 &5.29 & 70.8 &6.08 \\
& +DPCO & 79.2 &7.13 & 70.2 &7.37 & 91.2 &6.85 & 58.7 &3.94 & 62.1 &5.50 & 72.3 &6.16 \\
 & +RCO (Ours) & \textbf{90.4} &\textbf{7.28} & \textbf{77.4} &\textbf{7.59} & \textbf{94.3} &\textbf{7.28} & \textbf{70.7} &\textbf{4.45} & \textbf{72.5} &\textbf{5.86} & \textbf{81.1} &\textbf{6.49} \\ \midrule
\multirow{3}{*}{\begin{tabular}[c]{@{}c@{}}LLaMA-2-\\13B-Chat\end{tabular}}    & Base model & 78.4 &7.06 & 74.7 &7.56 & 86.9 &6.64 & 66.3 &4.21 & 61.1 &5.39 & 73.5 &6.17 \\
& +DPCO & 76.5 &6.91 & 79.9 &7.56 & 86.0 &6.51 & 56.1 &3.68 & 69.2 &5.70 & 73.5 &6.08 \\
& +RCO (Ours) & \textbf{86.9} &\textbf{7.20} & \textbf{85.3} &\textbf{7.60} & \textbf{96.2} &\textbf{7.22} & \textbf{71.0} &\textbf{4.38} & \textbf{70.1} &5.69 & \textbf{81.9} &\textbf{6.42} \\ \midrule
\multirow{3}{*}{\begin{tabular}[c]{@{}c@{}}LLaMA-3-\\8B-Instruct\end{tabular}} & Base model & 75.8 &7.02 & 78.4 &7.60 & 82.0 &6.55 & 69.8 &4.41 & 73.1 &6.08 & 75.8 &6.33 \\
& +DPCO & 68.7 &6.71 & 79.0 &7.30 & 82.4 &6.76 & 63.8 &4.26 & 70.5 &5.32 & 72.9 &6.07 \\
& +RCO (Ours) & \textbf{87.0} &\textbf{7.17} & \textbf{86.0 }&\textbf{7.74} & \textbf{94.2} &\textbf{7.03} & \textbf{76.2} &\textbf{4.84} & \textbf{78.3} &\textbf{6.33} & \textbf{84.3} &\textbf{6.62} \\ \midrule
\multirow{3}{*}{Auto-J-13B} & Base model & 60.0 &6.83 & 85.6 &7.60 & 79.1 &6.50 & 60.4 &3.85 & 63.1 &5.22 & 69.7 &6.00 \\
& +DPCO & 68.3 &6.99 & 80.3 &7.47 & 87.9 &6.57 & 64.5 &4.06 & 70.2 &5.73 & 74.2 &6.16 \\
& +RCO (Ours) & \textbf{77.5} &\textbf{7.09} & \textbf{83.7} &\textbf{7.54} & \textbf{91.9} &\textbf{7.17} & \textbf{68.5} &\textbf{4.38} & \textbf{72.3} &5.63 & \textbf{78.8} &\textbf{6.36} \\ \midrule
\multirow{3}{*}{\begin{tabular}[c]{@{}c@{}}UltraCM-\\13B\end{tabular}}         & Base model & 53.9 &6.50 & 72.6 &7.24 & 63.8 &5.88 & 65.2 &4.27 & 62.4 &5.25 & 63.6 &5.83 \\
& +DPCO & 69.8 &6.92 & 85.1 &7.61 & 84.8 &6.47 & 62.4 &3.98 & 67.0 &5.33 & 73.8 &6.06 \\
& +RCO (Ours) & \textbf{74.7 }&\textbf{7.05} & \textbf{89.6} &\textbf{7.66 }& \textbf{88.4 }&\textbf{6.79 }& \textbf{72.0 }&\textbf{4.50} & \textbf{69.3 }&\textbf{5.46 }&\textbf{ 78.8} & \textbf{6.29} \\ \midrule
\end{tabular}
}
\caption{Evaluation results of our method and baselines, in terms of critique utility (CU) and refinement quality score (RQS). Summ. and QA. are the shorter form of summarization and question answering tasks, respectively. The critique utility results reported in the table are multiplied by 100.}
\label{tab:main}
\end{table*}

\begin{table*}[ht]
\resizebox{1.0\textwidth}{!}{
\begin{tabular}{cccccccc|cc|cc|c}
\toprule
\multirow{2}{*}{\begin{tabular}[c]{@{}c@{}}Base \\ Model\end{tabular}} & \multirow{2}{*}{Method} & \multicolumn{6}{c}{QA.} & \multicolumn{2}{c}{Math} & \multicolumn{2}{c}{Code} & Reward \\ \cmidrule{3-13} 
& & BBH & \begin{tabular}[c]{@{}c@{}}GPQA-\\Diamond\end{tabular} & MMLU & \begin{tabular}[c]{@{}c@{}}MMLU-\\Pro\end{tabular} & \begin{tabular}[c]{@{}c@{}}MMLU-\\Redux\end{tabular} & Truthful & MATH & GSM-8K & MBPP & MBPP+ & \begin{tabular}[c]{@{}c@{}}Reward-\\Bench\end{tabular} \\ \midrule
\multicolumn{2}{c}{Initial Answer} & 39.67 & 27.65 & 52.48 & 33.14 & 48.79 & 49.12 & 17.10 & 42.05 & 35.84 & 32.85 & --\\ \midrule
\multicolumn{13}{c}{BASELINES}  \\ \midrule
\multicolumn{2}{c}{LLaMA-2-70B-Chat} & 44.19 & 26.77 & 60.02 & 37.59 & 53.55 & 50.23 & 21.75 & 52.50 & 39.00 & 36.72 & 65.9\\
\multicolumn{2}{c}{LLaMA-3-70B-Instruct} & 58.01 & 30.30 & 67.39 & 46.36 & 61.24 & 58.21 & 36.90 & 76.14 & 41.60 & 39.26 & 77.1 \\
\multicolumn{2}{c}{Self-refinement} & 43.65 & 29.60 & 54.79 & 35.49 & 49.52 & 52.39 & 22.91 & 54.42 & 37.20 & 36.50 & -- \\
\multicolumn{2}{c}{Aligner} & 42.39 & 24.15 & 52.56 & 32.39 & 48.59 & 46.61 & 15.62 & 42.74 & 30.88 & 30.54 & -- \\ \midrule
\multicolumn{13}{c}{OUR METHOD}\\ \midrule
\multirow{3}{*}{\begin{tabular}[c]{@{}c@{}}LLaMA-2-\\7B-Chat\end{tabular}}     & Base model & 38.22 & 27.47 & 56.96 & 35.29 & 50.82 & 48.32 & 19.65 & 48.92 & 36.16 & 34.74 & 59.5 \\
& +DPCO & 45.52 & 28.79 & 60.46 & 37.22 & 53.81 & 50.43 & 20.50 & 55.31 & 36.44 & 35.12 & 55.8 \\
 & +RCO (Ours) & \textbf{48.06} & \textbf{28.83} & \textbf{63.17} & \textbf{42.30} & \textbf{57.67} & \textbf{54.96} & \textbf{24.20} & \textbf{64.31} & \textbf{39.72} & \textbf{38.42} & \textbf{67.1} \\ \midrule
\multirow{3}{*}{\begin{tabular}[c]{@{}c@{}}LLaMA-2-\\13B-Chat\end{tabular}}    & Base model & 40.30 & 26.49 & 57.33 & 34.93 & 50.23 & 49.81 & 18.13 & 44.90 & 38.24 & 35.05 & 62.0\\
& +DPCO & 43.41 & 27.17 & 59.42 & 36.52 & 53.04 & 50.28 & 16.00 & 57.21 & 38.56 & 35.82 & 64.9 \\
& +RCO (Ours) & \textbf{47.89} & \textbf{28.35} & \textbf{62.48} & \textbf{43.01} & \textbf{57.98} & \textbf{55.31} & \textbf{23.60} & \textbf{65.54} & \textbf{39.92} & \textbf{37.69} & \textbf{68.3} \\ \midrule
\multirow{3}{*}{\begin{tabular}[c]{@{}c@{}}LLaMA-3-\\8B-Instruct\end{tabular}} & Base model & 45.33 & 25.47 & 59.41 & 37.88 & 53.24 & 54.32 & 25.36 & 58.11 & 39.32 & 37.73 & 69.7 \\
& +DPCO & 45.37 & 25.76 & 59.21 & 36.69 & 53.52 & 50.70 & 21.91 & 60.79 & 39.04 & 39.63 & 65.6 \\
& +RCO (Ours) & \textbf{53.51} & \textbf{30.06} & \textbf{67.74} & \textbf{46.23} & \textbf{61.30} & \textbf{59.58} & \textbf{32.31} & \textbf{70.97} & \textbf{41.08} & \textbf{40.34} & \textbf{77.6} \\ \midrule
\multirow{3}{*}{Auto-J-13B} & Base model & 37.75 & 23.56 & 52.90 & 28.65 & 46.63 & 50.40 & 16.28 & 41.93 & 34.07 & 32.86 & 61.9 \\
& +DPCO & 43.22 & 26.61 & 58.47 & 36.21 & 53.88 & 54.72 & 19.95 & 54.68 & 34.93 & 34.01 & 64.2 \\
& +RCO (Ours) & \textbf{46.18} & \textbf{28.04} & \textbf{60.37} & \textbf{40.89} & \textbf{56.47} & \textbf{56.19} & \textbf{26.52} & \textbf{63.92} & \textbf{38.80} & \textbf{37.23} & \textbf{67.3} \\ \midrule
\multirow{3}{*}{\begin{tabular}[c]{@{}c@{}}UltraCM-\\13B\end{tabular}}         & Base model & 45.93 & 25.96 & 57.19 & 33.73 & 50.90 & 55.98 & 20.85 & 56.27 & 35.40 & 34.46& 67.8 \\
& +DPCO & 47.29 & 26.35 & 60.03 & 35.98 & 55.94 & 54.31 & 22.40 & 58.93 & 36.85 & 35.27& 71.0 \\
& +RCO (Ours) & \textbf{49.75} & \textbf{27.83} & \textbf{64.42} & \textbf{40.74} & \textbf{57.89} & \textbf{55.62} & \textbf{25.17} & \textbf{63.45} & \textbf{38.92} & \textbf{37.74} & \textbf{71.9} \\ \midrule
\end{tabular}
}
\caption{Accuracy results on multiple established benchmarks and RewardBench.}
\label{tab:refine}
\end{table*}

\begin{figure*}[ht]
    \centering
    \subfigure[Critique evaluation results]{
        \centering
        \includegraphics[width=0.48\textwidth]{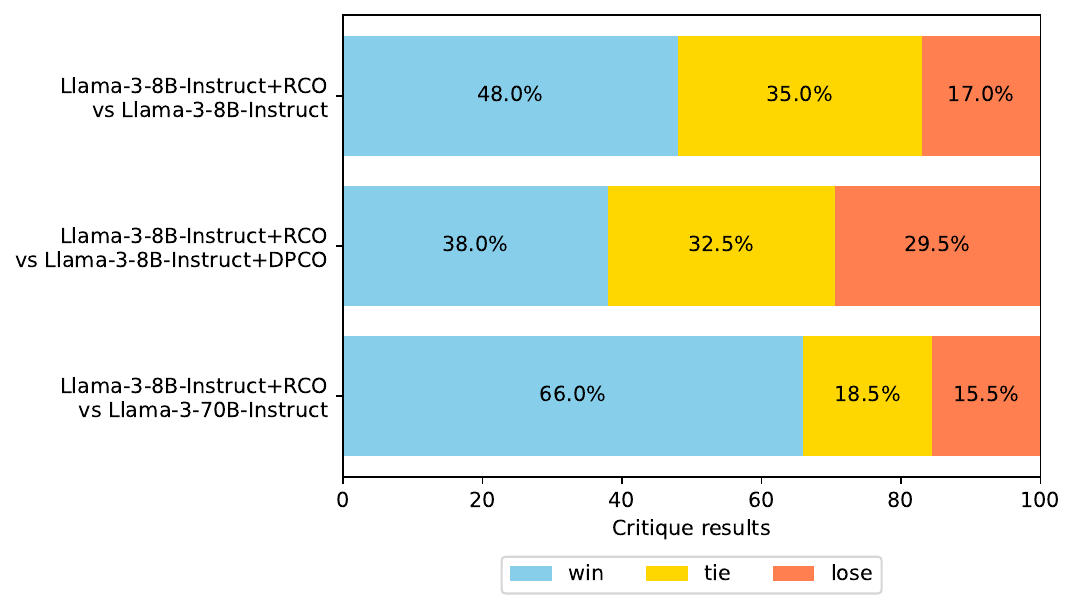}
        \label{sf:hc}
    }
    \subfigure[Refinement evaluation results]{
        \centering
        \includegraphics[width=0.48\textwidth]{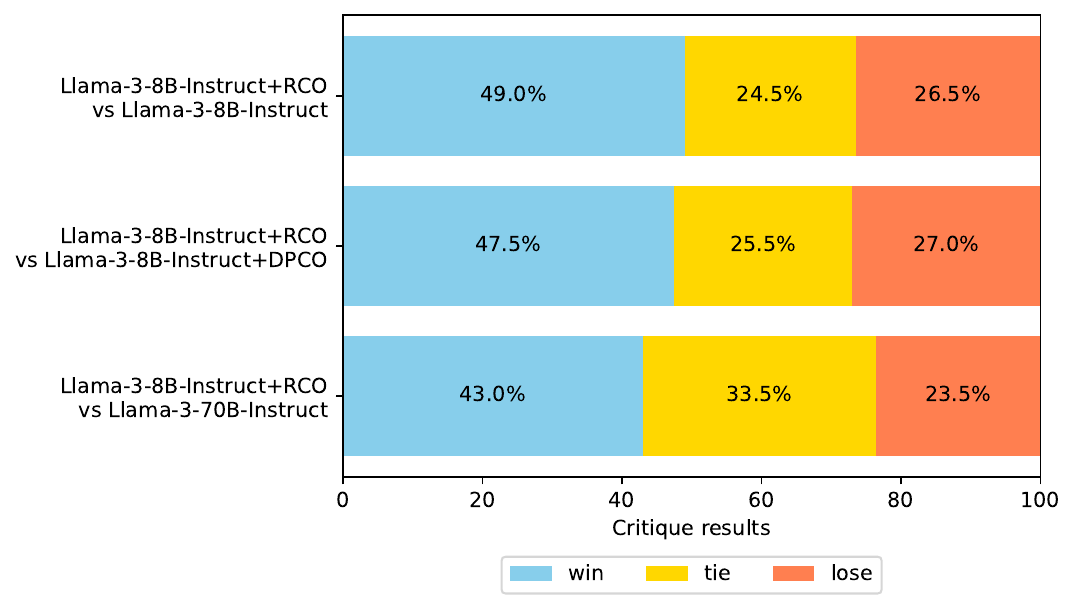}
        \label{sf:hr}
    }
    \caption{Visualization of human evaluation results in terms of critique quality evaluation and refinement quality evaluation.}
    \label{fig:human}
\end{figure*}

\begin{table*}[ht]
\resizebox{1.0\textwidth}{!}{
\begin{tabular}{ccccccc|cccc}
\toprule
\multirow{2}{*}{RCO (Ours)}                                                                             & \multirow{2}{*}{VS.}                                                                             & \multirow{2}{*}{Task} & \multicolumn{4}{c|}{Critique}              & \multicolumn{4}{c}{Refinement}             \\ \cmidrule{4-11} 
&   &   & Win (\%) & Tie (\%) & Loss (\%) & Gap (\%) & Win (\%) & Tie (\%) & Loss (\%) & Gap (\%) \\ \midrule
\multirow{18}{*}{\begin{tabular}[c]{@{}c@{}}LLaMa-3-\\      8B-Instruct\\      +MSE\end{tabular}} & \multirow{6}{*}{\begin{tabular}[c]{@{}c@{}}LLaMa-3-\\      8B-Instruct\end{tabular}}             & Dialog                & 50.0     & 30.0     & 20.0      & +30.0    & 55.0     & 22.5     & 22.5      & +32.5     \\
&   & Summ.   & 60.0     & 25.0     & 15.0      & +45.0    & 57.5     & 17.5     & 25.0      & +32.5    \\
&   & QA.        & 42.5     & 35.0     & 22.5      & +20.0    & 37.5     & 20.0     & 42.5      & -5.0     \\
&   & Math      & 47.5     & 35.0     & 17.5      & +30.0    & 55.0     & 30.0     & 15.0      & +40.0    \\
&   & Code      & 40.0     & 50.0     & 10.0      & +30.0    & 40.0     & 32.5     & 27.5      & +12.5    \\
&   & Overall   & 48.0     & 35.0     & 17.0      & \textbf{+31.0}    & 49.0     & 24.5     & 26.5      & \textbf{+22.5}    \\ \cmidrule{2-11} 
&  \multirow{6}{*}{\begin{tabular}[c]{@{}c@{}}LLaMa-3-\\      8B-Instruct\\      +DPO\end{tabular}}          & Dialog                & 25.0     & 32.5     & 42.5      & -17.5    & 35.0     & 22.5     & 42.5 & -7.5     \\
&   & Summ.               & 52.5     & 15.0     & 32.5      & +20.0    & 30.0     & 50.0     & 20.0      & +10.0     \\
&   & QA.                    & 32.5     & 27.5     & 40.0      & -7.5     & 47.5     & 15.0     & 37.5      & +10.0    \\
&   & Math                  & 37.5     & 45.0     & 17.5      & +20.0    & 57.5     & 22.5     & 20.0      & +37.5    \\
&   & Code                  & 42.5     & 42.5     & 15.0      & +27.5    & 67.5     & 17.5     & 15.0      & +52.5    \\
&   & Overall               & 38.0     & 32.5     & 29.5      & \textbf{+8.5}     & 47.5     & 25.5     & 27.0      & \textbf{+20.5}    \\ \cmidrule{2-11} 
& \multirow{6}{*}{\begin{tabular}[c]{@{}c@{}}LLaMa-3-\\      70B-Instruct\end{tabular}}  & Dialog                & 80.0     & 12.5     & 7.5       & +72.5    & 40.0     & 32.5     & 27.5      & +12.5    \\
&   & Summ.               & 75.0     & 15.0     & 10.0      & +65.0    & 40.0     & 45.0     & 15.0      & +25.0    \\
&   & QA.                    & 75.0     & 17.5     & 7.5       & +67.5    & 62.5     & 15.0     & 22.5      & +40.0    \\
&   & Math                  & 50.0     & 10.0     & 40.0      & +10.0    & 42.5     & 32.5     & 25.0      & +17.5    \\
&   & Code                  & 50.0     & 37.5     & 12.5      & +37.5    & 30.0     & 42.5     & 27.5      & +2.5     \\
&   & Overall               & 66.0     & 18.5     & 15.5      & \textbf{+50.5}    & 43.0     & 33.5     & 23.5      & \textbf{+19.5}    \\ \bottomrule
\end{tabular}
}
\caption{Human evaluation results across different tasks.}
\label{tab:human}
\end{table*}

\section{Main Results}

\subsection{Critique Utility and Response Quality}
Our comprehensive evaluation demonstrates that RCO-trained critic models consistently outperform all baseline approaches across multiple dimensions. As shown in Table \ref{tab:main}, RCO achieves significant improvements in both CU and RQS metrics compared to baseline models. This performance advantage holds true across all base model architectures and sizes, validating the effectiveness of our training paradigm.

Specifically, RCO-trained critic models consistently outperform baselines across all model sizes and architectures. Notably, smaller RCO-trained models (e.g., \textit{LLaMA-2-7B-Chat}) surpass their larger counterparts (e.g., \textit{LLaMA-2-70B-Chat}), demonstrating strong parameter efficiency. The performance gains are particularly significant in mathematical reasoning and code generation tasks. In contrast, DPCO shows only marginal improvements over base models, indicating limited effectiveness in leveraging critique signals. While DPCO performs comparably in dialog and summarization tasks, RCO maintains a clear advantage in complex reasoning domains.


\subsection{Refined Response Accuracy Results}

Table \ref{tab:refine} presents comprehensive results across multiple benchmarks. Our analysis yields two important findings: (1) RCO consistently outperforms both base models and DPCO-trained variants across all benchmarks, including challenging datasets like GPQA-Diamond, MMLU-Pro, and MBPP-Plus. This demonstrates the method's robustness in generating actionable critiques that lead to verifiable improvements. (2) The performance advantage holds across different model scales, with RCO-trained smaller models frequently matching or exceeding the capabilities of much larger base models. This suggests our training paradigm effectively distills critique generation expertise.

\subsection{Pairwise Critique Ability Results}

The RewardBench evaluation (Table \ref{tab:refine}, rightmost column) reveals an emergent capability: RCO-trained models achieve superior performance on pairwise preference judgment despite no explicit training for this task. This suggests that: (1) Critique generation training may develop generalizable preference modeling abilities. (2) The RCO objective aligns well with the underlying mechanisms of reward modeling. (3) There exists potential synergy between critique generation and preference learning.

\subsection{Human Evaluation}

In the human evaluation, we focus on the top-performing base model \textit{LLaMA-3-8B-Instruct} for cost efficiency, comparing RCO against three baselines: its base version, \textit{LLaMA-3-70B-Instruct}, and DPCO with. Figure \ref{fig:human} shows RCO's consistent superiority in both critique and refinement quality.
These results underscore the efficacy of our approach in enhancing actor model responses through structured critique generation and refinement.

Analyzing human evaluation results across different tasks reported in Table \ref{tab:human} we conclude that RCO excels in mathematical reasoning and code generation versus base/DPCO models, and outperforms larger models in dialogue/summarization tasks. A detailed case analysis on the reason of these distinctions is presented in Section \ref{sec:case} and Appendix \ref{app:case}.

An intriguing observation from the human evaluation is that human preferences for critiques do not always align with the preferences for the refinements generated from those critiques. Specifically, while DPCO critiques are most preferred than the other two baselines, all baselines yield comparable refinement quality. This suggest that preferred critiques don't guarantee better refinements, highlighting the inherent complexity of the critique-refinement process.

\begin{figure*}[t]
    \centering
    \subfigure[Ours vs Base model]{
        \centering
        \includegraphics[width=0.27\textwidth]{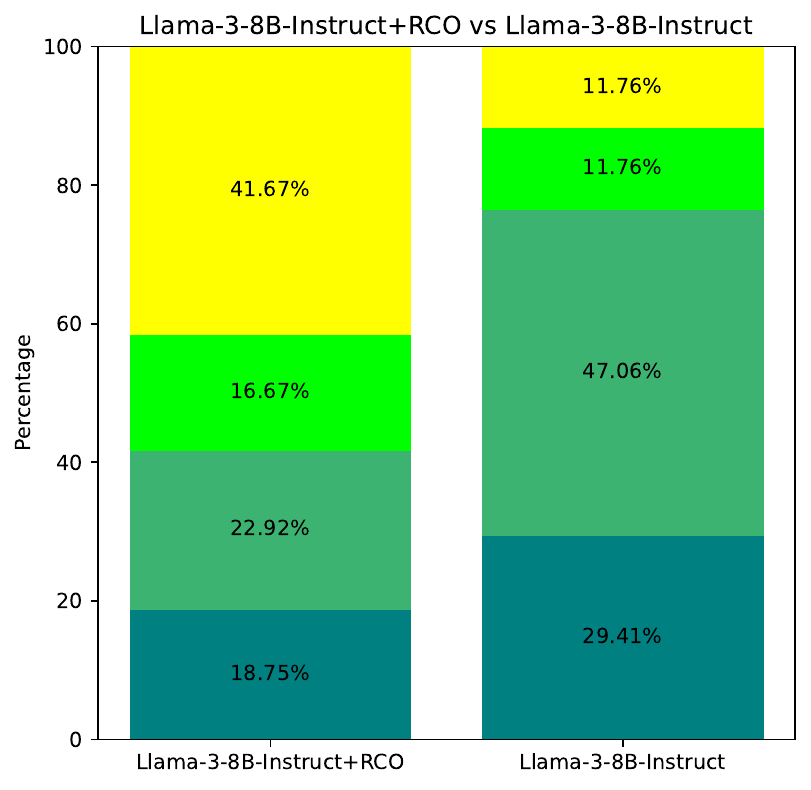}
        \label{sf:c0}
    }
    \subfigure[Ours vs DPCO]{
        \centering
        \includegraphics[width=0.27\textwidth]{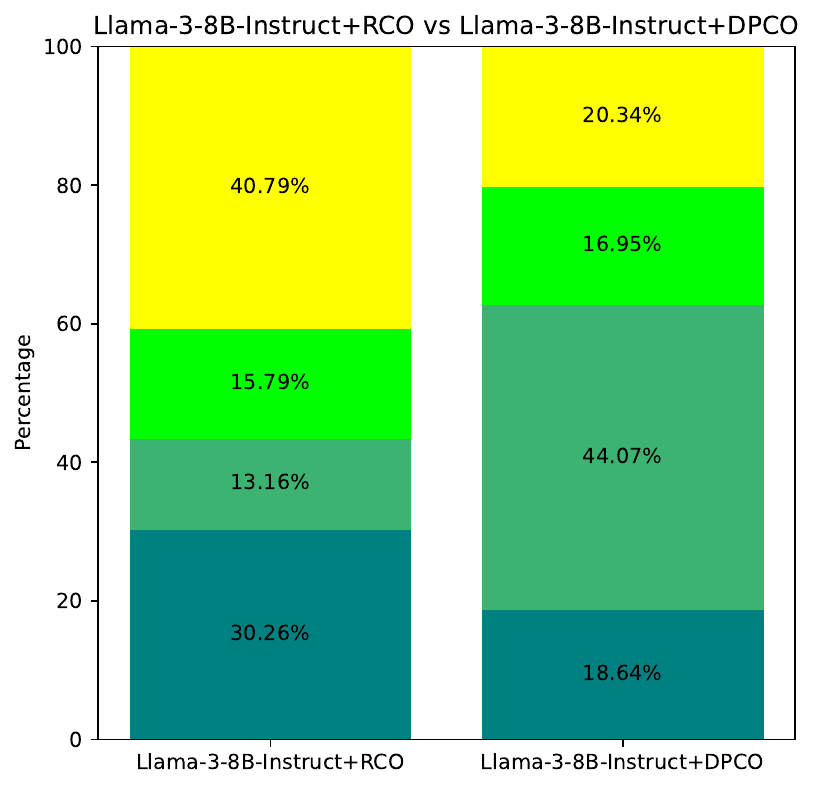}
        \label{sf:c1}
    }
    \subfigure[Ours vs Large model]{
        \centering
        \includegraphics[width=0.36\textwidth]{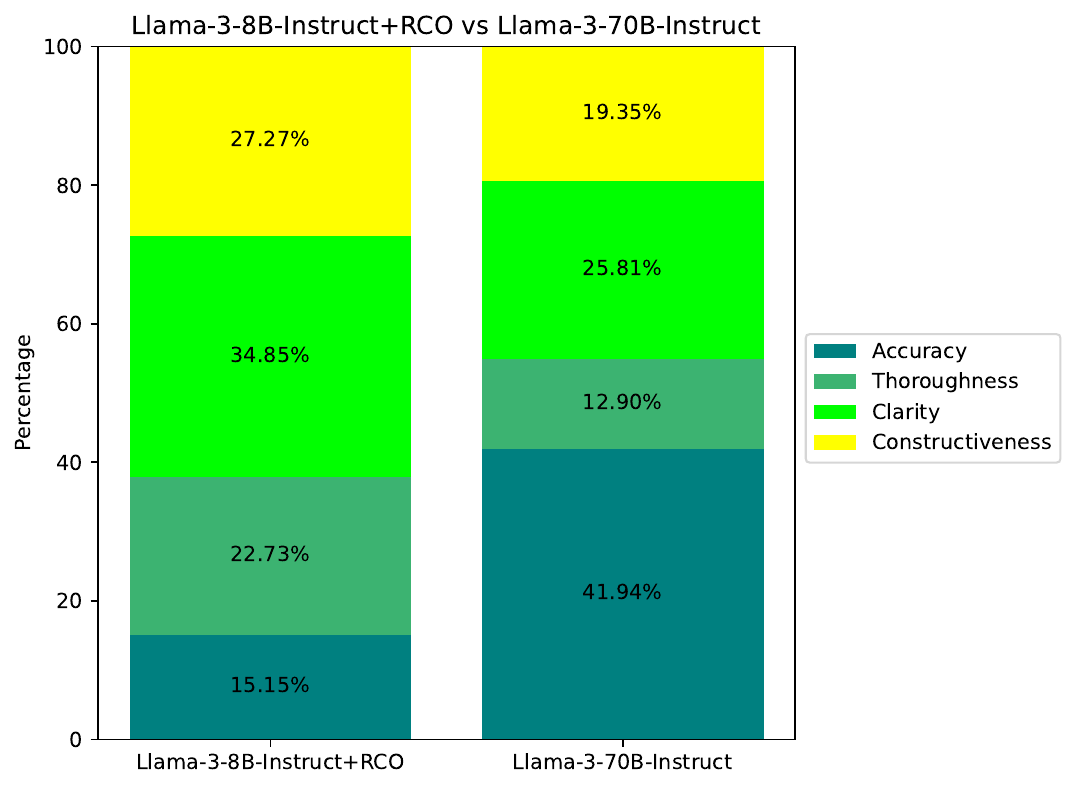}
        \label{sf:c2}
    }
    \caption{The percentage distribution of the reasons why each chosen critique is preferred in human evaluation.}
    \label{fig:case}
\end{figure*}

\section{Analysis}

\subsection{Weak-to-Strong Generalization Ability}

Recent studies in LLMs have highlighted the growing trend of scalable oversight~\cite{burns2023weak}, which requires smaller models to provide meaningful feedback to enhance the performance of larger models. To evaluate whether RCO facilitates weak-to-strong generalization, we reanalyze the experimental results presented in Table \ref{tab:main}, displaying the CU and RPS results with respect to actor models generating the initial responses in the test dataset.

Based on the findings in Table \ref{tab:actor}, the following key observations can be made:
(1) Both CU and RQS improve as the actor model increases in size and capability (from LLaMA-2 to LLaMA-3), primarily because the stronger models exhibit a greater ability to interpret critiques and follow refinement instructions.
(2) When compared to self-refinement (refinement without critique), the RCO-trained models show superior performance, particularly for initial responses generated by \textit{LLaMA-3-70B-Instruct}. In contrast, DPCO-trained models perform worse than self-refinement in this case.
(3) A comparison between RCO, base models, and DPCO reveals that RCO significantly enhances the performance of larger models (70B) on refined responses. For smaller actor models, however, RCO only slightly outperforms DPCO.
These findings suggest that RCO enables improved weak-to-strong generalization over DPCO and other baseline methods by supervising critique training with the outcome of refinement.

\subsection{Scaling Preference Judge Model for RCO}

The training process of RCO depends on the preference judgment of both refined and initial responses. As such, the accuracy of preference judgment plays a crucial role in the overall performance of RCO. In our experiment, we leverage \textit{Qwen2.5-72B-Instruct} (Qwen) to evaluate the preferences of responses. To assess the impact of the selection of preference judge model, we investigate four distinct variations: 
(1) \textbf{Stronger LLMs}, such as \textit{GPT-4o}~\cite{achiam2023gpt}; (2) \textbf{Open-source LLM Critic Models}, such as \textit{Skywork-Critic-LLaMA-3.1-8B} (Skywork)~\cite{skyworkcritic2024}; (3) \textbf{Reward Models}, such as \textit{InternLM2-7B-Reward} (InternLM)~\cite{cai2024internlm2} and PairRM~\cite{jiang2023llm}; and (4) \textbf{Self-rewarding} method~\cite{yuan2024self}, in which the base critic model evaluates the preferences for refinements under its own outputs. We conduct experiments using the \textit{LLaMA-3-8B-Instruct} as the base critic model and report the results of RCO with different preference judge models, as shown in Table \ref{tab:annotator}. 
Additionally, we provide the accuracy of preferences for each judge model using the established pairwise judgment benchmark, RewardBench~\cite{lambert2024rewardbench}. 

Our findings indicate that PairRM, which has the lowest pairwise accuracy on RewardBench, results in the worst performance. However, despite achieving the best pairwise accuracy, the InternLM model does not yield the top scores in both CU and RQS. Overall, the performances of RCO training with judge models except PairRM are quite similar, with CU scores ranging from 82\% to 84\% and RQS scores between 6.55 and 6.65. 
Notably, the self-rewarding method produces performance on par with that of stronger preference judge models, underscoring the potential of a self-supervised training paradigm.

\subsection{Scaling Numbers of Critiques and Refinements}

In the data construction phase, we establish $N=4$ critiques per initial answer to ensure reliable estimation of the partition function $Z_\beta(y_0,x)$, and $M=5$ refinements per critique for accurate Critique Utility (CU) estimation. This configuration generates $2\times N\times M$ pairwise comparisons per training sample, presenting substantial computational demands. To investigate cost-reduction opportunities, we systematically evaluate the impact of decreasing $N$ and $M$ values on model performance. As shown in Figure \ref{fig:nm}, we report the CU and RQS results of two base models \textit{LLaMA-2-7B-Chat} and \textit{LLaMA-3-8B-Instruct} over theoretically minimal settings ($N=2$, $M=1$). 
Our findings reveals that (1) CU, RQS and accuracy metrics exhibit slight but consistent degradation as $N$ or $M$ decreases, confirming that higher sample sizes contribute to training effectiveness. (2) Remarkably, RCO maintains significant superiority over DPCO across all tested configurations, demonstrating the inherent advantages of our approach regardless of sampling parameters.

\subsection{Why RCO Generate Effective Critiques for Refinement}
\label{sec:case}

To further explore why our method produces more effective critiques for refining actor models, we conduct an analysis asking human evaluators to select the primary reason for their preference of a given critique from the following four options: ``The critique is more accurate and correct (Accuracy)'', ``The critique provides a more thorough analysis of errors in the initial response (Thoroughness)'', ``The critique is clearer and better structured (Clarity)'' and ``The critique offers constructive suggestions or provides detailed steps for revision (Constructiveness)''.
The percentage distribution of reasons of critique preferences for each evaluated model (our method, the base model, the large model, and DPCO) is shown in Figure \ref{fig:case}. Additionally, we conduct a case study with five representative cases from the human evaluation dataset, including critiques generated by our method and baseline approaches, as well as the refinements produced by \textit{LLaMA-2-7B-Chat}, in Figures \ref{fig:case1}-\ref{fig:case5} of Appendix \ref{app:case} due to space constraints.

The results in Figure \ref{fig:case} clearly demonstrate that our method excels in generating critiques that are not only correct, clear, and well-structured, but also offer constructive and feasible suggestions or detailed revision steps that are easy for the actor model to follow. 
In contrast, DPCO, which rely on LLM-annotated preferences of critique pairs, often produce critiques that are richer and more thorough in their analysis but tend to offer vague or less specific suggestions. Furthermore, DPCO occasionally generate incorrect critiques or misidentify the target of critique. These findings underscore the effectiveness of our method in training critic models that generate precise, helpful critiques, significantly enhancing the iterative improvement of actor models.

\section{Conclusion}

In this paper we introduced RCO, a novel approach for training critic models to improve actor model refinement. 
By proposing a supervision scheme based on preferences for refined responses, we eliminate the need for direct critique quality assessment while rewarding critiques that drive meaningful improvements. 
Rigorous evaluations across five tasks show significant advancements in both critique quality and refinement capabilities compared to existing methods. 
Our in-depth analysis underscores the effectiveness of RCO and its potential to enhance the alignment between critic and actor models. 
These contributions provide valuable insights for designing more efficient and scalable systems for model refinement, laying the foundation for future research in this field.

\section*{Limitation}

Despite the effectiveness and strong potential of RCO for broader applications, several limitations warrant further investigation and improvement. 
One key limitation is the inaccurate estimation of critique utility \( \text{CU}(c_i|y_0,x) \) and regularization term \( Z_{\beta}(y_0,x) \). In our study, we sample 4 critiques to estimate \( Z_{\beta}(y_0,x) \) and 5 refined responses to estimate \( \text{CU}(c_i|y_0,x) \), which maybe not sufficient. However, to achieve more accurate estimates, more data and preference judgments are required, which increase the cost for data collection.
Additionally, our approach focus solely on critic models, failing to train actor models for improved utilization of critiques for refinement.
Moving forward, we aim to develop more efficient methods for training the critic model. Furthermore, we are interested in advancing techniques for actor models to better interpret natural language critiques and leverage them to enhance their responses.

\section*{Ethical Consideration}

In this work, we leveraged several available datasets to construct the training and test dataset of RCO. The HH-RLHF~\cite{bai2022training}, TL;DR~\cite{stiennon2020learning}, Commonsense QA~\cite{talmor-etal-2019-commonsenseqa}, MATH~\cite{hendrycks2021measuring}, GSM8K~\cite{cobbe2021training}, TheoremQA~\cite{chen2023theoremqa}, TabMWP~\cite{lu2022dynamic}, BBH~\cite{suzgun2022challenging}, GPQA~\cite{rein2024gpqa}, MMLU~\cite{hendrycks2020measuring} and MBPP-Plus~\cite{cassano2023multipl} are under MIT licenses; the CNN DailyMail~\cite{see2017get}, MathQA~\cite{amini-etal-2019-mathqa}, AQuA~\cite{ling2017program}, PiQA~\cite{bisk2020piqa}, MBPP~\cite{austin2021program}, HumanEval~\cite{zheng2023codegeex}, Chatbot Arena~\cite{chiang2024chatbot}, MMLU-Pro~\cite{wang2024mmlu}, TruthfulQA~\cite{lin2021truthfulqa}, CriticEval~\cite{lan2024criticbench} and RewardBench~\cite{lambert2024rewardbench} are under Apache licenses; the AmbigQA~\cite{min2020ambigqa}, ARC-Challenge~\cite{allenai:arc}, DS-1000~\cite{Lai2022DS1000} and MMLU-Redux~\cite{gema2024we} are under CC BY-SA licenses; the NYT~\cite{sandhaus2008new} dataset is under LDC license; the ELI5~\cite{fan2019eli5} dataset is under BSD license.

In these datasets, there exists some instructions with security issues. However, in RCO training, we constructed optimized prompt pairs that provide safety enhancements to these unsafe instructions, further mitigating the security issues.

\section*{Acknowledgements}
This work was supported by the National Science Foundation for Distinguished Young Scholars (with No. 62125604). We would also like to thank China Telecom Research INstitute and Zhipu AI for sponsoring the computational resources in this work.

\bibliography{custom}

\begin{thebibliography}{68}
\providecommand{\natexlab}[1]{#1}

\bibitem[{Achiam et~al.(2023)Achiam, Adler, Agarwal, Ahmad, Akkaya, Aleman, Almeida, Altenschmidt, Altman, Anadkat et~al.}]{achiam2023gpt}
Josh Achiam, Steven Adler, Sandhini Agarwal, Lama Ahmad, Ilge Akkaya, Florencia~Leoni Aleman, Diogo Almeida, Janko Altenschmidt, Sam Altman, Shyamal Anadkat, et~al. 2023.
\newblock Gpt-4 technical report.
\newblock \emph{arXiv preprint arXiv:2303.08774}.

\bibitem[{Aky{\"u}rek et~al.(2023)Aky{\"u}rek, Aky{\"u}rek, Madaan, Kalyan, Clark, Wijaya, and Tandon}]{akyurek2023rl4f}
Afra~Feyza Aky{\"u}rek, Ekin Aky{\"u}rek, Aman Madaan, Ashwin Kalyan, Peter Clark, Derry Wijaya, and Niket Tandon. 2023.
\newblock Rl4f: Generating natural language feedback with reinforcement learning for repairing model outputs.
\newblock \emph{arXiv preprint arXiv:2305.08844}.

\bibitem[{Amini et~al.(2019)Amini, Gabriel, Lin, Koncel-Kedziorski, Choi, and Hajishirzi}]{amini-etal-2019-mathqa}
Aida Amini, Saadia Gabriel, Shanchuan Lin, Rik Koncel-Kedziorski, Yejin Choi, and Hannaneh Hajishirzi. 2019.
\newblock \href {https://doi.org/10.18653/v1/N19-1245} {{M}ath{QA}: Towards interpretable math word problem solving with operation-based formalisms}.
\newblock In \emph{Proceedings of the 2019 Conference of the North {A}merican Chapter of the Association for Computational Linguistics: Human Language Technologies, Volume 1 (Long and Short Papers)}, pages 2357--2367, Minneapolis, Minnesota. Association for Computational Linguistics.

\bibitem[{Austin et~al.(2021)Austin, Odena, Nye, Bosma, Michalewski, Dohan, Jiang, Cai, Terry, Le et~al.}]{austin2021program}
Jacob Austin, Augustus Odena, Maxwell Nye, Maarten Bosma, Henryk Michalewski, David Dohan, Ellen Jiang, Carrie Cai, Michael Terry, Quoc Le, et~al. 2021.
\newblock Program synthesis with large language models.
\newblock \emph{arXiv preprint arXiv:2108.07732}.

\bibitem[{Bai et~al.(2022)Bai, Jones, Ndousse, Askell, Chen, DasSarma, Drain, Fort, Ganguli, Henighan et~al.}]{bai2022training}
Yuntao Bai, Andy Jones, Kamal Ndousse, Amanda Askell, Anna Chen, Nova DasSarma, Dawn Drain, Stanislav Fort, Deep Ganguli, Tom Henighan, et~al. 2022.
\newblock Training a helpful and harmless assistant with reinforcement learning from human feedback.
\newblock \emph{arXiv preprint arXiv:2204.05862}.

\bibitem[{Ben~Allal et~al.(2022)Ben~Allal, Muennighoff, Kumar~Umapathi, Lipkin, and von Werra}]{bigcode-evaluation-harness}
Loubna Ben~Allal, Niklas Muennighoff, Logesh Kumar~Umapathi, Ben Lipkin, and Leandro von Werra. 2022.
\newblock A framework for the evaluation of code generation models.
\newblock \url{https://github.com/bigcode-project/bigcode-evaluation-harness}.

\bibitem[{Bisk et~al.(2020)Bisk, Zellers, Gao, Choi et~al.}]{bisk2020piqa}
Yonatan Bisk, Rowan Zellers, Jianfeng Gao, Yejin Choi, et~al. 2020.
\newblock Piqa: Reasoning about physical commonsense in natural language.
\newblock In \emph{Proceedings of the AAAI conference on artificial intelligence}, pages 7432--7439.

\bibitem[{Burns et~al.(2023)Burns, Izmailov, Kirchner, Baker, Gao, Aschenbrenner, Chen, Ecoffet, Joglekar, Leike et~al.}]{burns2023weak}
Collin Burns, Pavel Izmailov, Jan~Hendrik Kirchner, Bowen Baker, Leo Gao, Leopold Aschenbrenner, Yining Chen, Adrien Ecoffet, Manas Joglekar, Jan Leike, et~al. 2023.
\newblock Weak-to-strong generalization: Eliciting strong capabilities with weak supervision.
\newblock \emph{arXiv preprint arXiv:2312.09390}.

\bibitem[{Cai et~al.(2024)Cai, Cao, Chen, Chen, Chen, Chen, Chen, Chen, Chen, Chu et~al.}]{cai2024internlm2}
Zheng Cai, Maosong Cao, Haojiong Chen, Kai Chen, Keyu Chen, Xin Chen, Xun Chen, Zehui Chen, Zhi Chen, Pei Chu, et~al. 2024.
\newblock Internlm2 technical report.
\newblock \emph{arXiv preprint arXiv:2403.17297}.

\bibitem[{Cao et~al.(2024)Cao, Lam, Duan, Liu, Zhang, and Chen}]{cao2024compassjudger}
Maosong Cao, Alexander Lam, Haodong Duan, Hongwei Liu, Songyang Zhang, and Kai Chen. 2024.
\newblock Compassjudger-1: All-in-one judge model helps model evaluation and evolution.
\newblock \emph{arXiv preprint arXiv:2410.16256}.

\bibitem[{Cassano et~al.(2023)Cassano, Gouwar, Nguyen, Nguyen, Phipps-Costin, Pinckney, Yee, Zi, Anderson, Feldman et~al.}]{cassano2023multipl}
Federico Cassano, John Gouwar, Daniel Nguyen, Sydney Nguyen, Luna Phipps-Costin, Donald Pinckney, Ming-Ho Yee, Yangtian Zi, Carolyn~Jane Anderson, Molly~Q Feldman, et~al. 2023.
\newblock Multipl-e: a scalable and polyglot approach to benchmarking neural code generation.
\newblock \emph{IEEE Transactions on Software Engineering}, 49(7):3675--3691.

\bibitem[{Chen et~al.(2024)Chen, Prasad, Saha, Stengel-Eskin, and Bansal}]{chen2024magicore}
Justin Chih-Yao Chen, Archiki Prasad, Swarnadeep Saha, Elias Stengel-Eskin, and Mohit Bansal. 2024.
\newblock Magicore: Multi-agent, iterative, coarse-to-fine refinement for reasoning.
\newblock \emph{arXiv preprint arXiv:2409.12147}.

\bibitem[{Chen et~al.(2023)Chen, Yin, Ku, Lu, Wan, Ma, Xu, Wang, and Xia}]{chen2023theoremqa}
Wenhu Chen, Ming Yin, Max Ku, Pan Lu, Yixin Wan, Xueguang Ma, Jianyu Xu, Xinyi Wang, and Tony Xia. 2023.
\newblock Theoremqa: A theorem-driven question answering dataset.
\newblock In \emph{Proceedings of the 2023 Conference on Empirical Methods in Natural Language Processing}, pages 7889--7901.

\bibitem[{Chiang et~al.(2024)Chiang, Zheng, Sheng, Angelopoulos, Li, Li, Zhang, Zhu, Jordan, Gonzalez et~al.}]{chiang2024chatbot}
Wei-Lin Chiang, Lianmin Zheng, Ying Sheng, Anastasios~Nikolas Angelopoulos, Tianle Li, Dacheng Li, Hao Zhang, Banghua Zhu, Michael Jordan, Joseph~E Gonzalez, et~al. 2024.
\newblock Chatbot arena: An open platform for evaluating llms by human preference.
\newblock \emph{arXiv preprint arXiv:2403.04132}.

\bibitem[{Clark et~al.(2018)Clark, Cowhey, Etzioni, Khot, Sabharwal, Schoenick, and Tafjord}]{allenai:arc}
Peter Clark, Isaac Cowhey, Oren Etzioni, Tushar Khot, Ashish Sabharwal, Carissa Schoenick, and Oyvind Tafjord. 2018.
\newblock Think you have solved question answering? try arc, the ai2 reasoning challenge.
\newblock \emph{arXiv:1803.05457v1}.

\bibitem[{Cobbe et~al.(2021)Cobbe, Kosaraju, Bavarian, Chen, Jun, Kaiser, Plappert, Tworek, Hilton, Nakano et~al.}]{cobbe2021training}
Karl Cobbe, Vineet Kosaraju, Mohammad Bavarian, Mark Chen, Heewoo Jun, Lukasz Kaiser, Matthias Plappert, Jerry Tworek, Jacob Hilton, Reiichiro Nakano, et~al. 2021.
\newblock Training verifiers to solve math word problems.
\newblock \emph{arXiv preprint arXiv:2110.14168}.

\bibitem[{Cui et~al.(2024)Cui, Yuan, Ding, Yao, He, Zhu, Ni, Xie, Xie, Lin, Liu, and Sun}]{cui2023ultrafeedback}
Ganqu Cui, Lifan Yuan, Ning Ding, Guanming Yao, Bingxiang He, Wei Zhu, Yuan Ni, Guotong Xie, Ruobing Xie, Yankai Lin, Zhiyuan Liu, and Maosong Sun. 2024.
\newblock {ULTRAFEEDBACK:} boosting language models with scaled {AI} feedback.
\newblock In \emph{Forty-first International Conference on Machine Learning}.

\bibitem[{Dubey et~al.(2024)Dubey, Jauhri, Pandey, Kadian, Al-Dahle, Letman, Mathur, Schelten, Yang, Fan et~al.}]{dubey2024llama}
Abhimanyu Dubey, Abhinav Jauhri, Abhinav Pandey, Abhishek Kadian, Ahmad Al-Dahle, Aiesha Letman, Akhil Mathur, Alan Schelten, Amy Yang, Angela Fan, et~al. 2024.
\newblock The llama 3 herd of models.
\newblock \emph{arXiv preprint arXiv:2407.21783}.

\bibitem[{Fan et~al.(2019)Fan, Jernite, Perez, Grangier, Weston, and Auli}]{fan2019eli5}
Angela Fan, Yacine Jernite, Ethan Perez, David Grangier, Jason Weston, and Michael Auli. 2019.
\newblock Eli5: Long form question answering.
\newblock \emph{arXiv preprint arXiv:1907.09190}.

\bibitem[{Gema et~al.(2024)Gema, Leang, Hong, Devoto, Mancino, Saxena, He, Zhao, Du, Madani et~al.}]{gema2024we}
Aryo~Pradipta Gema, Joshua Ong~Jun Leang, Giwon Hong, Alessio Devoto, Alberto Carlo~Maria Mancino, Rohit Saxena, Xuanli He, Yu~Zhao, Xiaotang Du, Mohammad Reza~Ghasemi Madani, et~al. 2024.
\newblock Are we done with mmlu?
\newblock \emph{arXiv preprint arXiv:2406.04127}.

\bibitem[{Hendrycks et~al.(2020)Hendrycks, Burns, Basart, Zou, Mazeika, Song, and Steinhardt}]{hendrycks2020measuring}
Dan Hendrycks, Collin Burns, Steven Basart, Andy Zou, Mantas Mazeika, Dawn Song, and Jacob Steinhardt. 2020.
\newblock Measuring massive multitask language understanding.
\newblock \emph{arXiv preprint arXiv:2009.03300}.

\bibitem[{Hendrycks et~al.(2021)Hendrycks, Burns, Kadavath, Arora, Basart, Tang, Song, and Steinhardt}]{hendrycks2021measuring}
Dan Hendrycks, Collin Burns, Saurav Kadavath, Akul Arora, Steven Basart, Eric Tang, Dawn Song, and Jacob Steinhardt. 2021.
\newblock Measuring mathematical problem solving with the math dataset.
\newblock \emph{arXiv preprint arXiv:2103.03874}.

\bibitem[{Huang et~al.(2023)Huang, Chen, Mishra, Zheng, Yu, Song, and Zhou}]{huang2023large}
Jie Huang, Xinyun Chen, Swaroop Mishra, Huaixiu~Steven Zheng, Adams~Wei Yu, Xinying Song, and Denny Zhou. 2023.
\newblock Large language models cannot self-correct reasoning yet.
\newblock \emph{arXiv preprint arXiv:2310.01798}.

\bibitem[{Ji et~al.(2024)Ji, Chen, Lou, Hong, Zhang, Pan, Dai, and Yang}]{ji2024aligner}
Jiaming Ji, Boyuan Chen, Hantao Lou, Donghai Hong, Borong Zhang, Xuehai Pan, Juntao Dai, and Yaodong Yang. 2024.
\newblock Aligner: Achieving efficient alignment through weak-to-strong correction.
\newblock \emph{arXiv preprint arXiv:2402.02416}.

\bibitem[{Jiang et~al.(2023)Jiang, Ren, and Lin}]{jiang2023llm}
Dongfu Jiang, Xiang Ren, and Bill~Yuchen Lin. 2023.
\newblock Llm-blender: Ensembling large language models with pairwise ranking and generative fusion.
\newblock In \emph{Proceedings of the 61st Annual Meeting of the Association for Computational Linguistics (Volume 1: Long Papers)}, pages 14165--14178.

\bibitem[{Jiang et~al.(2024)Jiang, Zhang, Weller, Weir, Van~Durme, and Khashabi}]{jiang2024self}
Dongwei Jiang, Jingyu Zhang, Orion Weller, Nathaniel Weir, Benjamin Van~Durme, and Daniel Khashabi. 2024.
\newblock Self-[in] correct: Llms struggle with refining self-generated responses.
\newblock \emph{arXiv preprint arXiv:2404.04298}.

\bibitem[{{Joshi} et~al.(2017){Joshi}, {Choi}, {Weld}, and {Zettlemoyer}}]{2017arXivtriviaqa}
Mandar {Joshi}, Eunsol {Choi}, Daniel {Weld}, and Luke {Zettlemoyer}. 2017.
\newblock \href {https://arxiv.org/abs/1705.03551} {{triviaqa: A Large Scale Distantly Supervised Challenge Dataset for Reading Comprehension}}.
\newblock \emph{arXiv e-prints}, arXiv:1705.03551.

\bibitem[{Ke et~al.(2024)Ke, Wen, Feng, Liu, Lei, Cheng, Wang, Zeng, Dong, Wang et~al.}]{ke2024critiquellm}
Pei Ke, Bosi Wen, Andrew Feng, Xiao Liu, Xuanyu Lei, Jiale Cheng, Shengyuan Wang, Aohan Zeng, Yuxiao Dong, Hongning Wang, et~al. 2024.
\newblock Critiquellm: Towards an informative critique generation model for evaluation of large language model generation.
\newblock In \emph{Proceedings of the 62nd Annual Meeting of the Association for Computational Linguistics (Volume 1: Long Papers)}, pages 13034--13054.

\bibitem[{Kim et~al.(2024)Kim, Suk, Longpre, Lin, Shin, Welleck, Neubig, Lee, Lee, and Seo}]{kim2024prometheus}
Seungone Kim, Juyoung Suk, Shayne Longpre, Bill~Yuchen Lin, Jamin Shin, Sean Welleck, Graham Neubig, Moontae Lee, Kyungjae Lee, and Minjoon Seo. 2024.
\newblock Prometheus 2: An open source language model specialized in evaluating other language models.
\newblock \emph{arXiv preprint arXiv:2405.01535}.

\bibitem[{Lai et~al.(2022)Lai, Li, Wang, Zhang, Zhong, Zettlemoyer, tau Yih, Fried, Wang, and Yu}]{Lai2022DS1000}
Yuhang Lai, Chengxi Li, Yiming Wang, Tianyi Zhang, Ruiqi Zhong, Luke Zettlemoyer, Scott~Wen tau Yih, Daniel Fried, Sida Wang, and Tao Yu. 2022.
\newblock Ds-1000: A natural and reliable benchmark for data science code generation.
\newblock \emph{ArXiv}, abs/2211.11501.

\bibitem[{Lambert et~al.(2024)Lambert, Pyatkin, Morrison, Miranda, Lin, Chandu, Dziri, Kumar, Zick, Choi et~al.}]{lambert2024rewardbench}
Nathan Lambert, Valentina Pyatkin, Jacob Morrison, LJ~Miranda, Bill~Yuchen Lin, Khyathi Chandu, Nouha Dziri, Sachin Kumar, Tom Zick, Yejin Choi, et~al. 2024.
\newblock Rewardbench: Evaluating reward models for language modeling.
\newblock \emph{arXiv preprint arXiv:2403.13787}.

\bibitem[{Lan et~al.(2024{\natexlab{a}})Lan, Zhang, Lyu, Li, Xu, Huang, Lin, Mao, and Chen}]{lan2024training}
Tian Lan, Wenwei Zhang, Chengqi Lyu, Shuaibin Li, Chen Xu, Heyan Huang, Dahua Lin, Xian-Ling Mao, and Kai Chen. 2024{\natexlab{a}}.
\newblock Training language models to critique with multi-agent feedback.
\newblock \emph{arXiv preprint arXiv:2410.15287}.

\bibitem[{Lan et~al.(2024{\natexlab{b}})Lan, Zhang, Xu, Huang, Lin, Chen, and Mao}]{lan2024criticbench}
Tian Lan, Wenwei Zhang, Chen Xu, Heyan Huang, Dahua Lin, Kai Chen, and Xian-ling Mao. 2024{\natexlab{b}}.
\newblock Criticbench: Evaluating large language models as critic.
\newblock \emph{arXiv preprint arXiv:2402.13764}.

\bibitem[{Li et~al.(2024{\natexlab{a}})Li, Jiang, Huang, Beigi, Zhao, Tan, Bhattacharjee, Jiang, Chen, Wu et~al.}]{li2024generation}
Dawei Li, Bohan Jiang, Liangjie Huang, Alimohammad Beigi, Chengshuai Zhao, Zhen Tan, Amrita Bhattacharjee, Yuxuan Jiang, Canyu Chen, Tianhao Wu, et~al. 2024{\natexlab{a}}.
\newblock From generation to judgment: Opportunities and challenges of llm-as-a-judge.
\newblock \emph{arXiv preprint arXiv:2411.16594}.

\bibitem[{Li et~al.(2024{\natexlab{b}})Li, Dong, Chen, Su, Zhou, Ai, Ye, and Liu}]{li2024llms}
Haitao Li, Qian Dong, Junjie Chen, Huixue Su, Yujia Zhou, Qingyao Ai, Ziyi Ye, and Yiqun Liu. 2024{\natexlab{b}}.
\newblock Llms-as-judges: a comprehensive survey on llm-based evaluation methods.
\newblock \emph{arXiv preprint arXiv:2412.05579}.

\bibitem[{Li et~al.(2023)Li, Sun, Yuan, Fan, Zhao, and Liu}]{li2023generative}
Junlong Li, Shichao Sun, Weizhe Yuan, Run-Ze Fan, Hai Zhao, and Pengfei Liu. 2023.
\newblock Generative judge for evaluating alignment.
\newblock \emph{arXiv preprint arXiv:2310.05470}.

\bibitem[{Lin et~al.(2021)Lin, Hilton, and Evans}]{lin2021truthfulqa}
Stephanie Lin, Jacob Hilton, and Owain Evans. 2021.
\newblock Truthfulqa: Measuring how models mimic human falsehoods.
\newblock \emph{arXiv preprint arXiv:2109.07958}.

\bibitem[{Lin et~al.(2024)Lin, Gou, Liang, Luo, Liu, and Yang}]{lin2024criticbench}
Zicheng Lin, Zhibin Gou, Tian Liang, Ruilin Luo, Haowei Liu, and Yujiu Yang. 2024.
\newblock Criticbench: Benchmarking llms for critique-correct reasoning.
\newblock \emph{arXiv preprint arXiv:2402.14809}.

\bibitem[{Ling et~al.(2017)Ling, Yogatama, Dyer, and Blunsom}]{ling2017program}
Wang Ling, Dani Yogatama, Chris Dyer, and Phil Blunsom. 2017.
\newblock Program induction by rationale generation: Learning to solve and explain algebraic word problems.
\newblock \emph{ACL}.

\bibitem[{Lu et~al.(2022)Lu, Qiu, Chang, Wu, Zhu, Rajpurohit, Clark, and Kalyan}]{lu2022dynamic}
Pan Lu, Liang Qiu, Kai-Wei Chang, Ying~Nian Wu, Song-Chun Zhu, Tanmay Rajpurohit, Peter Clark, and Ashwin Kalyan. 2022.
\newblock Dynamic prompt learning via policy gradient for semi-structured mathematical reasoning.
\newblock \emph{arXiv preprint arXiv:2209.14610}.

\bibitem[{Madaan et~al.(2023)Madaan, Tandon, Gupta, Hallinan, Gao, Wiegreffe, Alon, Dziri, Prabhumoye, Yang et~al.}]{madaan2023self}
Aman Madaan, Niket Tandon, Prakhar Gupta, Skyler Hallinan, Luyu Gao, Sarah Wiegreffe, Uri Alon, Nouha Dziri, Shrimai Prabhumoye, Yiming Yang, et~al. 2023.
\newblock Self-refine: Iterative refinement with self-feedback.
\newblock \emph{arXiv preprint arXiv:2303.17651}.

\bibitem[{McAleese et~al.(2024)McAleese, Pokorny, Uribe, Nitishinskaya, Trebacz, and Leike}]{mcaleese2024llm}
Nat McAleese, Rai~Michael Pokorny, Juan Felipe~Ceron Uribe, Evgenia Nitishinskaya, Maja Trebacz, and Jan Leike. 2024.
\newblock Llm critics help catch llm bugs.
\newblock \emph{arXiv preprint arXiv:2407.00215}.

\bibitem[{Min et~al.(2020)Min, Michael, Hajishirzi, and Zettlemoyer}]{min2020ambigqa}
Sewon Min, Julian Michael, Hannaneh Hajishirzi, and Luke Zettlemoyer. 2020.
\newblock Ambigqa: Answering ambiguous open-domain questions.
\newblock \emph{arXiv preprint arXiv:2004.10645}.

\bibitem[{Murugadoss et~al.(2024)Murugadoss, Poelitz, Drosos, Le, McKenna, Negreanu, Parnin, and Sarkar}]{murugadoss2024evaluating}
Bhuvanashree Murugadoss, Christian Poelitz, Ian Drosos, Vu~Le, Nick McKenna, Carina~Suzana Negreanu, Chris Parnin, and Advait Sarkar. 2024.
\newblock Evaluating the evaluator: Measuring llms' adherence to task evaluation instructions.
\newblock \emph{arXiv preprint arXiv:2408.08781}.

\bibitem[{Rafailov et~al.(2023)Rafailov, Sharma, Mitchell, Ermon, Manning, and Finn}]{rafailov2023direct}
Rafael Rafailov, Archit Sharma, Eric Mitchell, Stefano Ermon, Christopher~D Manning, and Chelsea Finn. 2023.
\newblock Direct preference optimization: Your language model is secretly a reward model.
\newblock \emph{arXiv preprint arXiv:2305.18290}.

\bibitem[{Rein et~al.(2024)Rein, Hou, Stickland, Petty, Pang, Dirani, Michael, and Bowman}]{rein2024gpqa}
David Rein, Betty~Li Hou, Asa~Cooper Stickland, Jackson Petty, Richard~Yuanzhe Pang, Julien Dirani, Julian Michael, and Samuel~R Bowman. 2024.
\newblock Gpqa: A graduate-level google-proof q\&a benchmark.
\newblock In \emph{First Conference on Language Modeling}.

\bibitem[{Richemond et~al.(2024)Richemond, Tang, Guo, Calandriello, Azar, Rafailov, Pires, Tarassov, Spangher, Ellsworth et~al.}]{richemond2024offline}
Pierre~Harvey Richemond, Yunhao Tang, Daniel Guo, Daniele Calandriello, Mohammad~Gheshlaghi Azar, Rafael Rafailov, Bernardo~Avila Pires, Eugene Tarassov, Lucas Spangher, Will Ellsworth, et~al. 2024.
\newblock Offline regularised reinforcement learning for large language models alignment.
\newblock \emph{arXiv preprint arXiv:2405.19107}.

\bibitem[{Sandhaus(2008)}]{sandhaus2008new}
Evan Sandhaus. 2008.
\newblock The new york times annotated corpus.
\newblock \emph{Linguistic Data Consortium, Philadelphia}, 6(12):e26752.

\bibitem[{Saunders et~al.(2022)Saunders, Yeh, Wu, Bills, Ouyang, Ward, and Leike}]{saunders2022self}
William Saunders, Catherine Yeh, Jeff Wu, Steven Bills, Long Ouyang, Jonathan Ward, and Jan Leike. 2022.
\newblock Self-critiquing models for assisting human evaluators.
\newblock \emph{arXiv preprint arXiv:2206.05802}.

\bibitem[{Scheurer et~al.(2023)Scheurer, Campos, Korbak, Chan, Chen, Cho, and Perez}]{scheurer2023training}
J{\'e}r{\'e}my Scheurer, Jon~Ander Campos, Tomasz Korbak, Jun~Shern Chan, Angelica Chen, Kyunghyun Cho, and Ethan Perez. 2023.
\newblock Training language models with language feedback at scale.
\newblock \emph{arXiv preprint arXiv:2303.16755}.

\bibitem[{See et~al.(2017)See, Liu, and Manning}]{see2017get}
Abigail See, Peter~J Liu, and Christopher~D Manning. 2017.
\newblock Get to the point: Summarization with pointer-generator networks.
\newblock \emph{arXiv preprint arXiv:1704.04368}.

\bibitem[{Shiwen et~al.(2024)Shiwen, Liang, Liu, Zeng, and Liu}]{skyworkcritic2024}
Tu~Shiwen, Zhao Liang, Chris~Yuhao Liu, Liang Zeng, and Yang Liu. 2024.
\newblock \href {https://huggingface.co/Skywork} {Skywork critic model series}.
\newblock \url{https://huggingface.co/Skywork}.

\bibitem[{Stiennon et~al.(2020)Stiennon, Ouyang, Wu, Ziegler, Lowe, Voss, Radford, Amodei, and Christiano}]{stiennon2020learning}
Nisan Stiennon, Long Ouyang, Jeffrey Wu, Daniel Ziegler, Ryan Lowe, Chelsea Voss, Alec Radford, Dario Amodei, and Paul~F Christiano. 2020.
\newblock Learning to summarize with human feedback.
\newblock \emph{Advances in Neural Information Processing Systems}, 33:3008--3021.

\bibitem[{Sun et~al.(2024)Sun, Li, Yuan, Yuan, Li, and Liu}]{sun2024critique}
Shichao Sun, Junlong Li, Weizhe Yuan, Ruifeng Yuan, Wenjie Li, and Pengfei Liu. 2024.
\newblock The critique of critique.
\newblock \emph{arXiv preprint arXiv:2401.04518}.

\bibitem[{Suzgun et~al.(2022)Suzgun, Scales, Sch{\"a}rli, Gehrmann, Tay, Chung, Chowdhery, Le, Chi, Zhou et~al.}]{suzgun2022challenging}
Mirac Suzgun, Nathan Scales, Nathanael Sch{\"a}rli, Sebastian Gehrmann, Yi~Tay, Hyung~Won Chung, Aakanksha Chowdhery, Quoc~V Le, Ed~H Chi, Denny Zhou, et~al. 2022.
\newblock Challenging big-bench tasks and whether chain-of-thought can solve them.
\newblock \emph{arXiv preprint arXiv:2210.09261}.

\bibitem[{Talmor et~al.(2019)Talmor, Herzig, Lourie, and Berant}]{talmor-etal-2019-commonsenseqa}
Alon Talmor, Jonathan Herzig, Nicholas Lourie, and Jonathan Berant. 2019.
\newblock \href {https://doi.org/10.18653/v1/N19-1421} {{C}ommonsense{QA}: A question answering challenge targeting commonsense knowledge}.
\newblock In \emph{Proceedings of the 2019 Conference of the North {A}merican Chapter of the Association for Computational Linguistics: Human Language Technologies, Volume 1 (Long and Short Papers)}, pages 4149--4158, Minneapolis, Minnesota. Association for Computational Linguistics.

\bibitem[{Team(2024)}]{qwen2.5}
Qwen Team. 2024.
\newblock \href {https://qwenlm.github.io/blog/qwen2.5/} {Qwen2.5: A party of foundation models}.

\bibitem[{Touvron et~al.(2023)Touvron, Martin, Stone, Albert, Almahairi, Babaei, Bashlykov, Batra, Bhargava, Bhosale et~al.}]{touvron2023llama}
Hugo Touvron, Louis Martin, Kevin Stone, Peter Albert, Amjad Almahairi, Yasmine Babaei, Nikolay Bashlykov, Soumya Batra, Prajjwal Bhargava, Shruti Bhosale, et~al. 2023.
\newblock Llama 2: Open foundation and fine-tuned chat models.
\newblock \emph{arXiv preprint arXiv:2307.09288}.

\bibitem[{Verga et~al.(2024)Verga, Hofstatter, Althammer, Su, Piktus, Arkhangorodsky, Xu, White, and Lewis}]{verga2024replacing}
Pat Verga, Sebastian Hofstatter, Sophia Althammer, Yixuan Su, Aleksandra Piktus, Arkady Arkhangorodsky, Minjie Xu, Naomi White, and Patrick Lewis. 2024.
\newblock Replacing judges with juries: Evaluating llm generations with a panel of diverse models.
\newblock \emph{arXiv preprint arXiv:2404.18796}.

\bibitem[{Wadhwa et~al.(2024)Wadhwa, Zhao, Li, and Durrett}]{wadhwa2024learning}
Manya Wadhwa, Xinyu Zhao, Junyi~Jessy Li, and Greg Durrett. 2024.
\newblock Learning to refine with fine-grained natural language feedback.
\newblock \emph{arXiv preprint arXiv:2407.02397}.

\bibitem[{Wang et~al.(2024{\natexlab{a}})Wang, Li, Chen, Cai, Zhu, Lin, Cao, Kong, Liu, Liu, and Sui}]{wang2024faireval}
Peiyi Wang, Lei Li, Liang Chen, Zefan Cai, Dawei Zhu, Binghuai Lin, Yunbo Cao, Lingpeng Kong, Qi~Liu, Tianyu Liu, and Zhifang Sui. 2024{\natexlab{a}}.
\newblock Large language models are not fair evaluators.
\newblock In \emph{Proceedings of the 62nd Annual Meeting of the Association for Computational Linguistics}, pages 9440--9450.

\bibitem[{Wang et~al.(2024{\natexlab{b}})Wang, Kulikov, Golovneva, Yu, Yuan, Dwivedi-Yu, Pang, Fazel-Zarandi, Weston, and Li}]{wang2024self}
Tianlu Wang, Ilia Kulikov, Olga Golovneva, Ping Yu, Weizhe Yuan, Jane Dwivedi-Yu, Richard~Yuanzhe Pang, Maryam Fazel-Zarandi, Jason Weston, and Xian Li. 2024{\natexlab{b}}.
\newblock Self-taught evaluators.
\newblock \emph{arXiv preprint arXiv:2408.02666}.

\bibitem[{Wang et~al.(2024{\natexlab{c}})Wang, Ma, Zhang, Ni, Chandra, Guo, Ren, Arulraj, He, Jiang et~al.}]{wang2024mmlu}
Yubo Wang, Xueguang Ma, Ge~Zhang, Yuansheng Ni, Abhranil Chandra, Shiguang Guo, Weiming Ren, Aaran Arulraj, Xuan He, Ziyan Jiang, et~al. 2024{\natexlab{c}}.
\newblock Mmlu-pro: A more robust and challenging multi-task language understanding benchmark.
\newblock In \emph{The Thirty-eight Conference on Neural Information Processing Systems Datasets and Benchmarks Track}.

\bibitem[{Wu et~al.(2024)Wu, Peng, Du, Zheng, Liu, Wu, Ma, Li, Yang, Zhou et~al.}]{wu2024comparative}
Siwei Wu, Zhongyuan Peng, Xinrun Du, Tuney Zheng, Minghao Liu, Jialong Wu, Jiachen Ma, Yizhi Li, Jian Yang, Wangchunshu Zhou, et~al. 2024.
\newblock A comparative study on reasoning patterns of openai's o1 model.
\newblock \emph{arXiv preprint arXiv:2410.13639}.

\bibitem[{Yuan et~al.(2024)Yuan, Pang, Cho, Sukhbaatar, Xu, and Weston}]{yuan2024self}
Weizhe Yuan, Richard~Yuanzhe Pang, Kyunghyun Cho, Sainbayar Sukhbaatar, Jing Xu, and Jason Weston. 2024.
\newblock Self-rewarding language models.
\newblock \emph{arXiv preprint arXiv:2401.10020}.

\bibitem[{Zheng et~al.(2023{\natexlab{a}})Zheng, Chiang, Sheng, Zhuang, Wu, Zhuang, Lin, Li, Li, Xing et~al.}]{zheng2023judging}
Lianmin Zheng, Wei-Lin Chiang, Ying Sheng, Siyuan Zhuang, Zhanghao Wu, Yonghao Zhuang, Zi~Lin, Zhuohan Li, Dacheng Li, Eric Xing, et~al. 2023{\natexlab{a}}.
\newblock Judging llm-as-a-judge with mt-bench and chatbot arena.
\newblock \emph{Advances in Neural Information Processing Systems}, 36:46595--46623.

\bibitem[{Zheng et~al.(2023{\natexlab{b}})Zheng, Xia, Zou, Dong, Wang, Xue, Wang, Shen, Wang, Li, Su, Yang, and Tang}]{zheng2023codegeex}
Qinkai Zheng, Xiao Xia, Xu~Zou, Yuxiao Dong, Shan Wang, Yufei Xue, Zihan Wang, Lei Shen, Andi Wang, Yang Li, Teng Su, Zhilin Yang, and Jie Tang. 2023{\natexlab{b}}.
\newblock Codegeex: A pre-trained model for code generation with multilingual benchmarking on humaneval-x.
\newblock In \emph{Proceedings of the 29th ACM SIGKDD Conference on Knowledge Discovery and Data Mining}, pages 5673--5684.

\bibitem[{Ziegler et~al.(2019)Ziegler, Stiennon, Wu, Brown, Radford, Amodei, Christiano, and Irving}]{ziegler2019fine}
Daniel~M Ziegler, Nisan Stiennon, Jeffrey Wu, Tom~B Brown, Alec Radford, Dario Amodei, Paul Christiano, and Geoffrey Irving. 2019.
\newblock Fine-tuning language models from human preferences.
\newblock \emph{arXiv preprint arXiv:1909.08593}.

\end{thebibliography}

\appendix

\begin{table*}[ht]
\resizebox{1.0\textwidth}{!}{
\begin{tabular}{cccc|cc|cc|cc|cc|cc}
\toprule
\multirow{2}{*}{\begin{tabular}[c]{@{}c@{}}Base   \\      Model\end{tabular}}    & \multirow{2}{*}{Method} & \multicolumn{2}{c}{LLaMA-2-7B} & \multicolumn{2}{c}{LLaMA-2-13B} & \multicolumn{2}{c}{LLaMA-2-70B} & \multicolumn{2}{c}{LLaMA-3-8B} & \multicolumn{2}{c}{LLaMA-3-70B} & \multicolumn{2}{c}{Overall}    \\ \cmidrule{3-14} 
& & CU & RQS & CU & RQS & CU & RQS & CU & RQS & CU & RQS & CU & RQS \\ \midrule
\multicolumn{2}{c}{Initial Answer} & -- & 4.84 & -- & 5.21 & -- & 4.93 & -- & 5.44 & -- & 5.82 & -- & 5.25\\ \midrule
\multicolumn{14}{c}{BASELINES}  \\ \midrule
\multicolumn{2}{c}{Self-refinement} & 58.7 &5.34 & 71.4 &6.09 & 72.2 &5.59 & 73.2 &6.08 & \underline{87.5} &\underline{6.78} & 72.6 &5.98 \\ 
\multicolumn{2}{c}{LLaMA-2-70B-Chat} & 69.1 &5.91 & 76.2 &6.28 & 71.4 &5.99 & 66.4 &6.32 & 79.1 &6.58 & 72.4 &6.21 \\
\multicolumn{2}{c}{LLaMA-3-70B-Instruct} & 78.9 &6.23 & 85.0 &6.64 & 85.6 &6.30 & 75.9 &6.35 & 85.5 &6.70 & 82.2 &6.44 \\
\midrule
\multicolumn{14}{c}{OUR METHOD}\\ \midrule
\multirow{3}{*}{\begin{tabular}[c]{@{}c@{}}LLaMA-2-\\7B-Chat\end{tabular}}     & Base model & 64.2 &5.63 & 73.9 &6.19 & 69.0 &5.94 & 65.2 &6.08 & 81.5 &6.54 & 70.8 &6.08 \\
& +DPCO & 68.2 &5.74 & 74.9 &6.20 & 70.2 &5.99 & 67.2 &6.18 & 80.8 &6.67 & 72.3 &6.16 \\
 & +RCO (Ours) & 70.0 &5.82 & 81.7 &6.51 & 80.5 &6.29 & 81.0 &6.62 & \underline{92.1} &\underline{7.21} & 81.1 &6.49 \\ \midrule
\multirow{3}{*}{\begin{tabular}[c]{@{}c@{}}LLaMA-2-\\13B-Chat\end{tabular}}    & Base model & 65.3 & 5.62 & 75.2 & 6.20 & 71.3 & 6.02 & 72.8 & 6.33 & 82.9 & 6.60 & 73.5 & 6.17 \\
& +DPCO & 66.5 &5.75 & 73.4 &6.23 & 73.8 &5.94 & 73.6 &6.03 & 80.4 &6.41 & 73.5 &6.07 \\
& +RCO (Ours) & 72.8 &5.74 & 81.4 &6.32 & 82.0 &6.14 & 79.7 &6.54 & \underline{93.6} &\underline{7.21} & 81.9 &6.42 \\ \midrule
\multirow{3}{*}{\begin{tabular}[c]{@{}c@{}}LLaMA-3-\\8B-Instruct\end{tabular}} & Base model & 72.1 &6.03 & 78.5 &6.55 & 78.2 &6.14 & 71.0 &6.38 & 79.4 &6.56 & 75.8 &6.33 \\
& +DPCO & 68.5 &5.83 & 74.1 &6.35 & 67.9 &5.80 & 71.9 &6.21 & 82.2 &6.64 & 72.9 &6.17 \\
& +RCO (Ours) & 82.6 &6.37 & 86.3 &6.75 & 85.2 &6.34 & 78.4 &6.69 & \underline{89.1} &\underline{6.95} & \textbf{84.3} &\textbf{6.62} \\ \midrule
\multirow{3}{*}{Auto-J-13B} & Base model & 60.1 &5.54 & 72.4 &6.14 & 69.3 &5.73 & 67.9 &6.05 & 78.5 &6.53 & 69.7 &6.00 \\
& +DPCO & 68.1 &5.77 & 74.1 &6.33 & 70.7 &5.93 & 74.2 &6.29 & 83.8 &6.49 & 74.2 &6.16 \\
& +RCO (Ours) & 68.5 &5.72 & 76.6 &6.43 & 76.1 &6.09 & 80.8 &6.56 & \underline{91.9} &\underline{7.00} & 78.8 & 6.36 \\ \midrule
\multirow{3}{*}{\begin{tabular}[c]{@{}c@{}}UltraCM-\\13B\end{tabular}} & Base model & 61.1 &5.63 & 63.1 &5.97 & 60.6 &5.51 & 61.6 &5.88 & 71.4 &6.14 & 63.6 &5.83 \\
& +DPCO & 66.4 &5.66 & 71.8 &6.10 & 70.5 &5.76 & 75.6 &6.16 & 84.8 &6.61 & 73.8 &6.06 \\
& +RCO (Ours) & 68.0 &5.75 & 76.2 &6.33 & 76.9 &6.07 & 82.6 & 6.45& \underline{90.3} &\underline{6.88} & 78.8 &6.29  \\ \midrule
\end{tabular}
}
\caption{Evaluation results of our method and baselines with respective to actor models that generate the initial response, in terms of critique utility (CU) and refinement quality score (RQS). We omit the suffixes (``-Chat'' and ``-Instruct'') of actor models (the first row) for better appearance. The critique utility results reported in the table are multiplied by 100.}
\label{tab:actor}
\end{table*}

\begin{table*}[ht]
\resizebox{1.0\textwidth}{!}{
\begin{tabular}{ccccc|cc|cc|cc|cc|cc}
\toprule
\multirow{2}{*}{\begin{tabular}[c]{@{}c@{}}Base \\ Model\end{tabular}} & \multirow{2}{*}{\begin{tabular}[c]{@{}c@{}}Preference \\ Judge\end{tabular}} & \multirow{2}{*}{\begin{tabular}[c]{@{}c@{}}Acc. @ \\ RewardBench\end{tabular}} & \multicolumn{2}{c}{Dialog} & \multicolumn{2}{c}{Summ.} & \multicolumn{2}{c}{QA.} & \multicolumn{2}{c}{Math} & \multicolumn{2}{c}{Code} & \multicolumn{2}{c}{Overall} \\ \cmidrule{4-15} 
& & & CU & RQS & CU & RQS & CU & RQS & CU & RQS & CU & RQS & CU & RQS \\ \midrule
\multirow{6}{*}{\begin{tabular}[c]{@{}c@{}}LLaMA-\\3-8B-\\Instruct\end{tabular}} & Qwen & 86.1\% & 87.0 & 7.17 & 86.0 & 7.74 & 94.2 & 7.03 & 76.2 & 4.84 & 78.3 & 6.33 & 84.3 & 6.62 \\
& GPT-4o & 84.6\% & 87.7 & 7.15 & 86.7 & 7.76 & 94.7 & 7.09 & 74.9 & 4.82 & 79.3 & 6.40 & 84.7 & 6.64 \\
& Skywork & 89.3\% & 83.3 & 7.08 & 86.3 & 7.69 & 91.2 & 7.02 & 73.2 & 4.76 & 73.7 & 6.22 & 83.9 & 6.55 \\
& InternLM & 90.1\% & 84.6 & 7.15 & 88.1 & 7.59 & 93.4 & 7.13 & 73.4 & 4.69 & 78.3 & 6.39 & 83.6 & 6.59 \\
& PairRM & 54.2\% & 79.9 & 7.04 & 82.9 & 7.57 & 83.8 & 6.49 & 69.8 & 4.58 & 71.2 & 6.20 & 77.5 & 6.38 \\
& Self-rewarding & 69.7\% & 85.3 & 7.13 & 84.1 & 7.65 & 92.3 & 7.27 & 74.3 & 4.76 & 76.9 & 6.30 & 82.6 & 6.62 \\
\midrule
\end{tabular}
}
\caption{Evaluation results of RCO training varying the preference judge model.}
\label{tab:annotator}
\end{table*}

\begin{table}[ht]
\resizebox{0.48\textwidth}{!}{
\begin{tabular}{c|c|c}
\toprule
Task                                                                              & Dataset        & Amount \\ \midrule
Dialog                                                                            & HH-RLHF~\cite{bai2022training}        & 2,000             \\ \midrule
\multirow{2}{*}{Summ.}                                                    & TL;DR~\cite{stiennon2020learning}          & 710               \\
 & CNN DailyMail~\cite{see2017get}  & 1,000             \\ \midrule
\multirow{5}{*}{QA.}     & Commonsense QA~\cite{talmor-etal-2019-commonsenseqa} & 500               \\
& Trivia QA~\cite{2017arXivtriviaqa}      & 500               \\
& AmbigQA~\cite{min2020ambigqa}        & 500               \\
& ARC-Challenge~\cite{allenai:arc}  & 500               \\
& ELI5~\cite{fan2019eli5}           & 500               \\ \midrule
\multirow{4}{*}{Math} & MathQA~\cite{amini-etal-2019-mathqa}         & 500               \\
& TheoremQA~\cite{chen2023theoremqa}      & 500               \\
& AQuA~\cite{ling2017program}           & 500               \\
& TabMWP~\cite{lu2022dynamic}         & 500               \\ \midrule
\multirow{2}{*}{Code}        & HumanEval~\cite{zheng2023codegeex}      & 820               \\
& DS-1000~\cite{Lai2022DS1000}        & 970               \\ \midrule
\multicolumn{2}{c|}{Total}                                                                          & 10,000      \\ \bottomrule
\end{tabular}
}
\caption{Statistics of all the 14 source datasets in the training dataset.}
\label{tab:dataset}
\end{table}

\begin{table}[ht]
\resizebox{0.48\textwidth}{!}{
\begin{tabular}{c|c|c}
\toprule
Task                                                                              & Dataset        & Amount \\ \midrule
Dialog  & Chatbot Arena~\cite{chiang2024chatbot}        & 500             \\ \midrule
Summ.  & New York Times~\cite{sandhaus2008new}          & 500 \\ \midrule
QA.     & PiQA~\cite{bisk2020piqa} & 500               \\ \midrule
\multirow{2}{*}{Math} & MATH~\cite{hendrycks2021measuring}         & 250               \\
& GSM8K~\cite{cobbe2021training}      & 250              \\ \midrule
Code   & MBPP~\cite{austin2021program}      & 500              \\ \midrule
\multicolumn{2}{c|}{Total}                                                                          & 2,500      \\ \bottomrule
\end{tabular}
}
\caption{Statistics of all the 7 source datasets in the test dataset.}
\label{tab:testset}
\end{table}

\begin{table}[ht]
\centering
\begin{tabular}{c|c|c|c}
\toprule
Evaluators Pair & (1, 2) & (1, 3) & (2, 3) \\ \midrule
Critique & 63\% & 58.5\% & 61.5\% \\
Refinement & 79.5\% & 73.5\% & 77\% \\ \bottomrule
\end{tabular}
\caption{Agreement rate between human evaluators pairs for both critique and refinement evaluation.}
\label{tab:agree}
\end{table}

\begin{table}[ht]
\centering
\begin{tabular}{c|c|c}
\toprule
Models & DPCO & RCO \\ \midrule
LLaMA-2-7B-Chat & 4.5h & 6.0h \\
LLaMA-2-13B-Chat & 13.4h & 16.9h \\
LLaMA-3-8B-Instruct & 7.1h & 8.8h \\
Auto-J-13B & 13.6h & 17h \\
UltraCM-13B & 13.2h & 16.4h \\ \bottomrule
\end{tabular}
\caption{Training time of DPCO and RCO.}
\label{tab:time}
\end{table}

\begin{figure*}[ht]
    \centering
    \subfigure[LLaMA-2-7B-Chat]{
        \centering
        \includegraphics[width=0.48\textwidth]{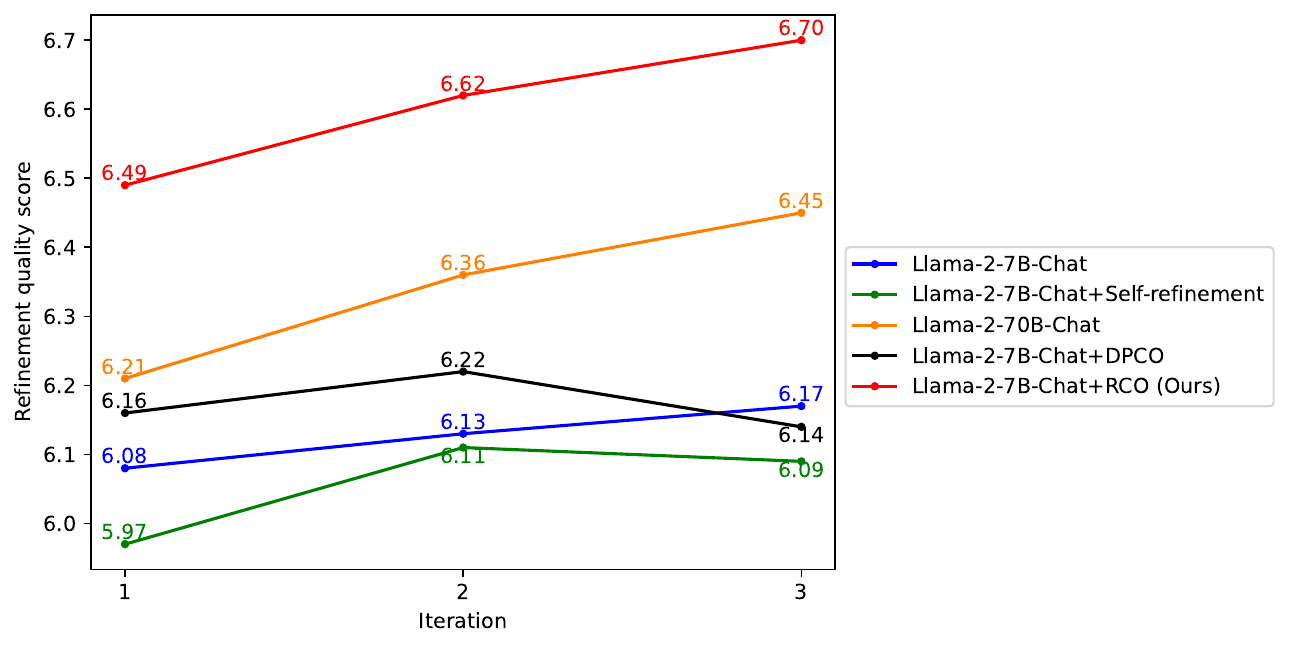}
        \label{sf:r7b}
    }
    \subfigure[LLaMA-3-8B-Instruct]{
        \centering
        \includegraphics[width=0.48\textwidth]{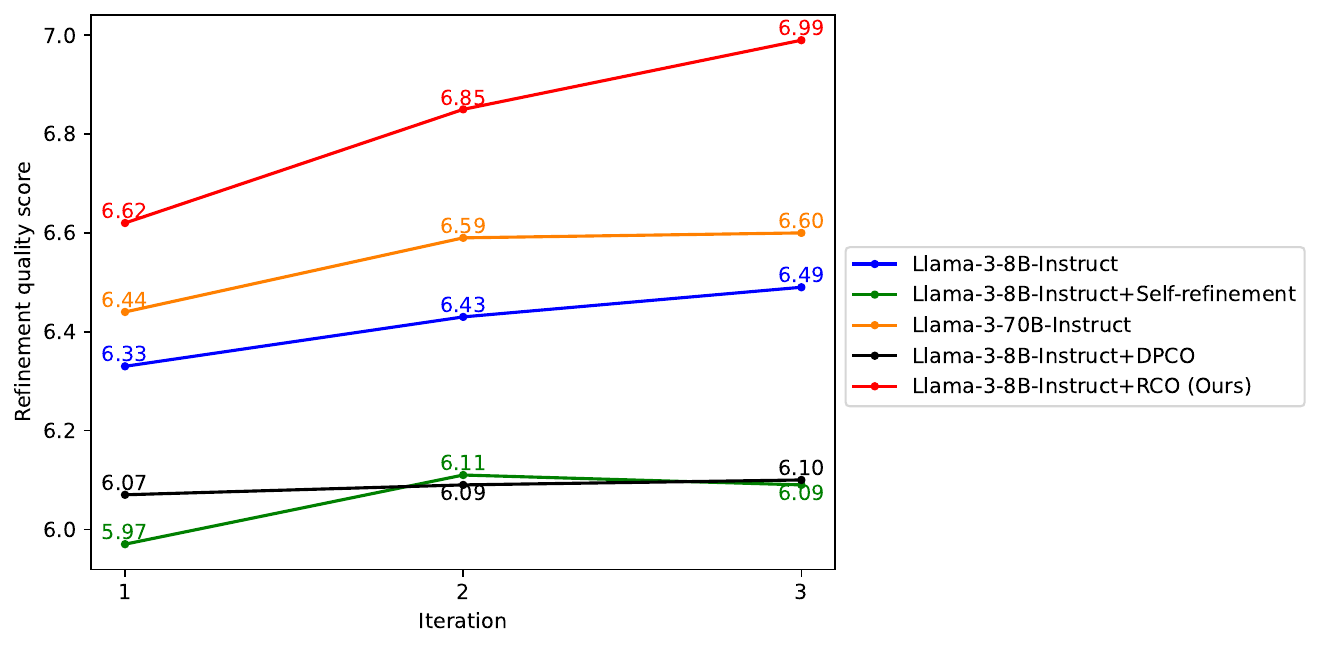}
        \label{sf:r8b}
    }
    \caption{Refinement quality score of iterative refinement results with different base models.}
    \label{fig:iter}
\end{figure*}

\begin{figure*}[t]
    \centering
    \subfigure[Llama-2-7B-Chat CU]{
        \centering
        \includegraphics[width=0.48\textwidth]{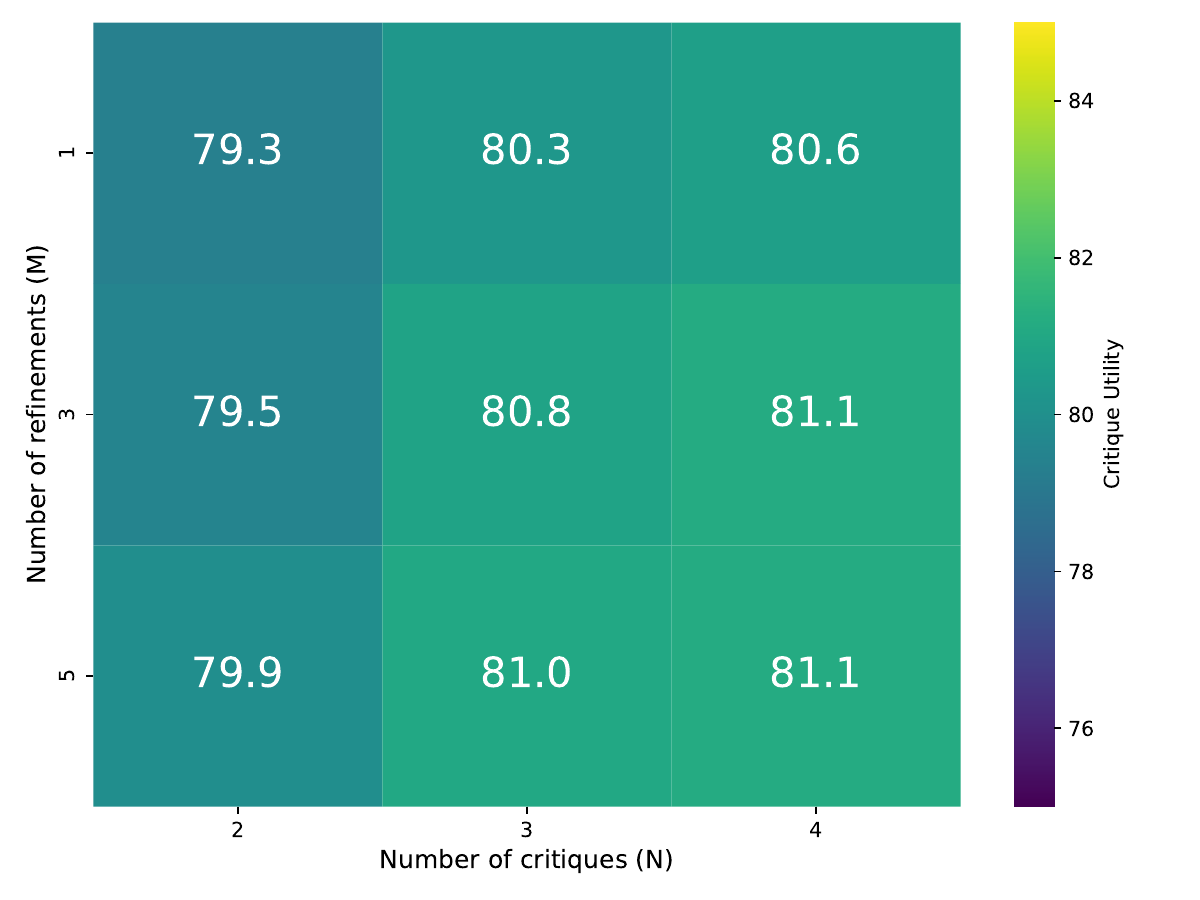}
        \label{sf:n0}
    }
    \subfigure[Llama-2-7B-Chat RQS]{
        \centering
        \includegraphics[width=0.48\textwidth]{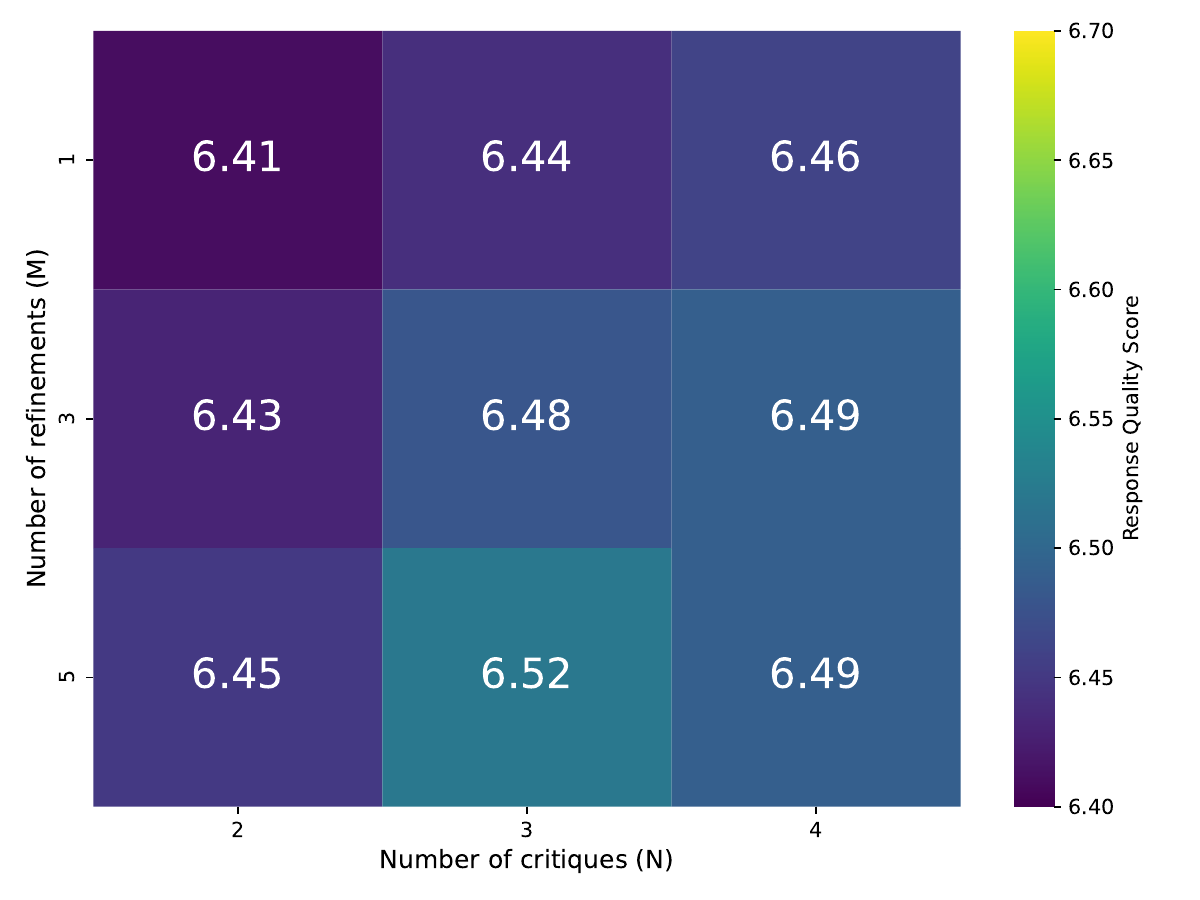}
        \label{sf:n1}
    }
    \subfigure[Llama-3-8B-Instruct CU]{
        \centering
        \includegraphics[width=0.48\textwidth]{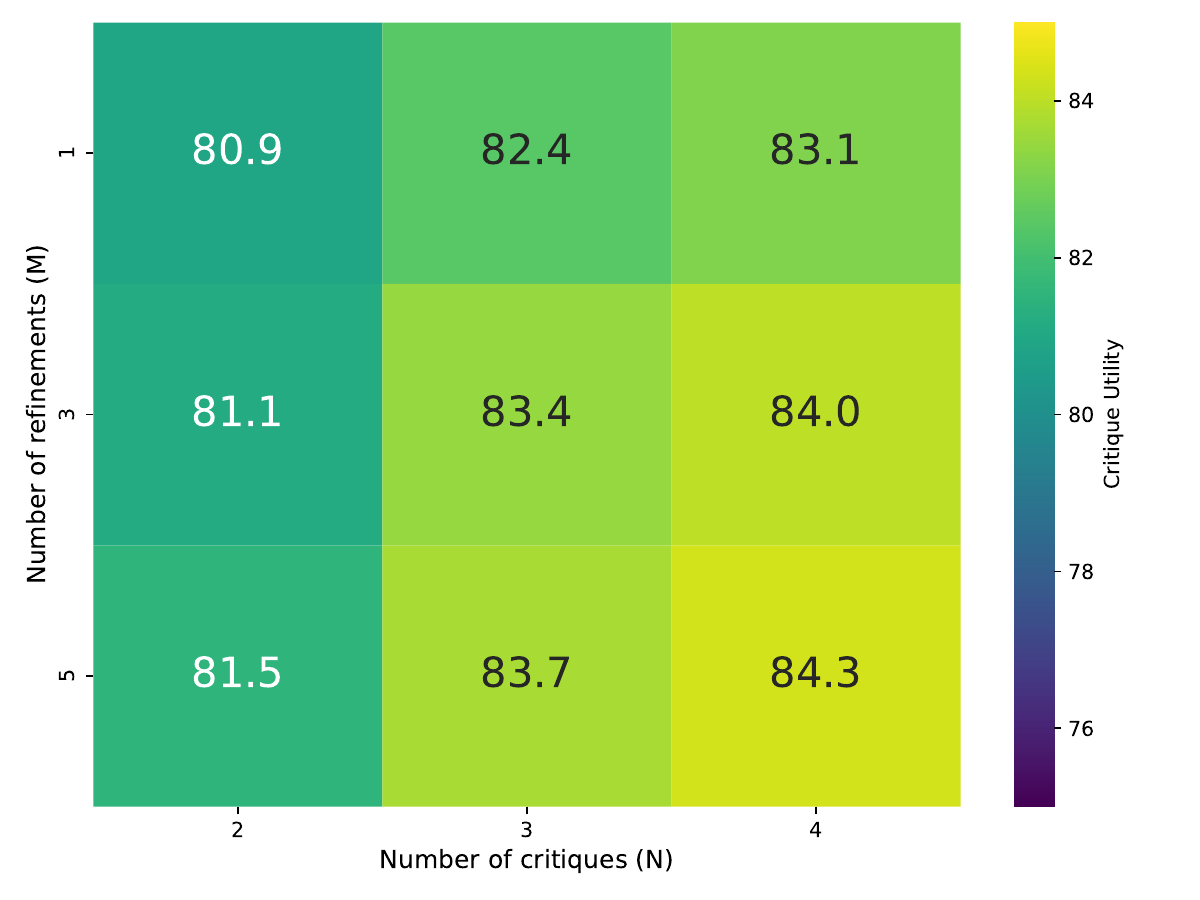}
        \label{sf:n2}
    }
    \subfigure[Llama-3-8B-Instruct RQS]{
        \centering
        \includegraphics[width=0.48\textwidth]{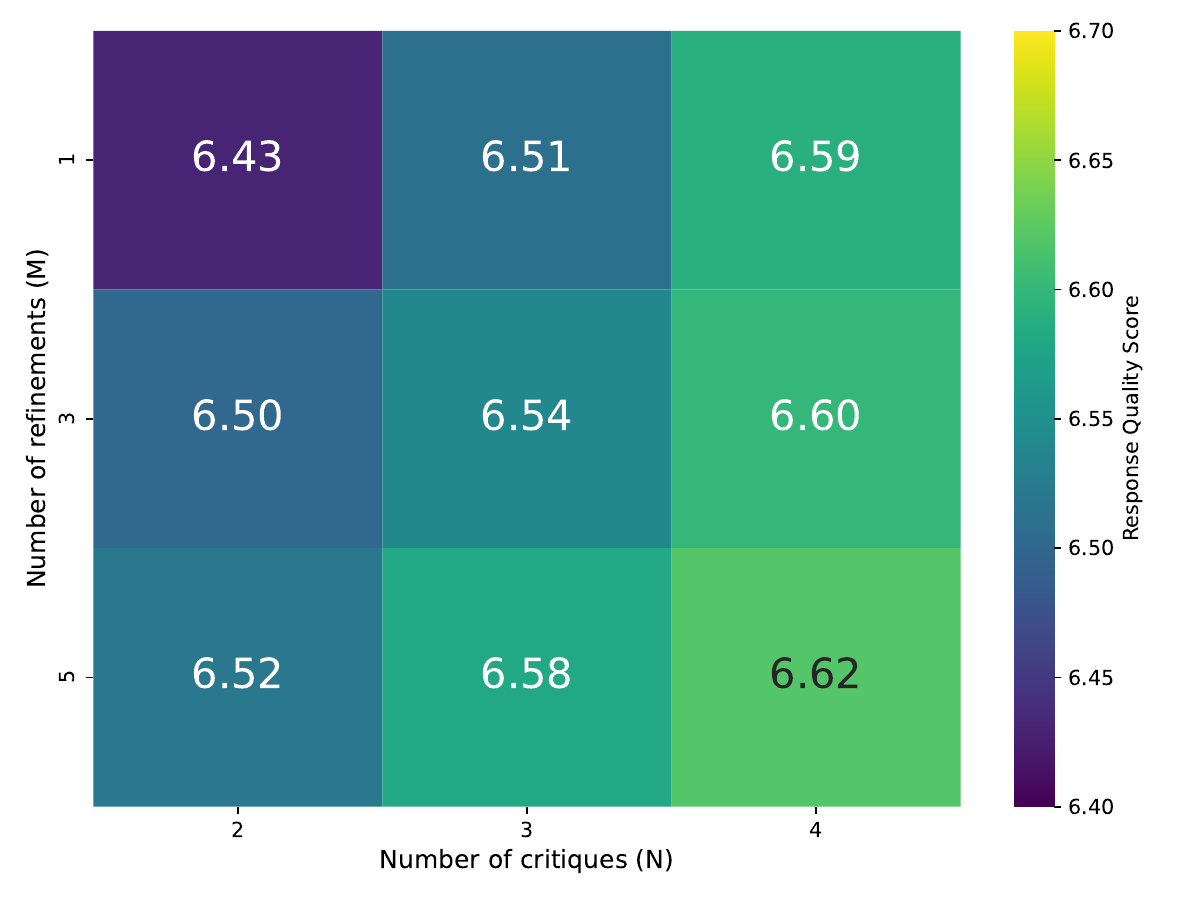}
        \label{sf:n2}
    }
    \caption{Heatmap results of CU and RQS metrics varying number of critiques ($N$) and number of refinements ($M$).}
    \label{fig:nm}
\end{figure*}

\section{Data Construction Prompts}
\label{app:dataprompt}

Since our dataset construction process involves multiple stages, including generation of initial responses, critiques, and refined responses, and 5 different tasks, we design distinct prompts for each stage and task. The prompts for initial response generation are shown in Table~\ref{tab:task0}. The prompts for critique collection are shown in Table \ref{tab:task11}-\ref{tab:task15}. The prompts for refinement generation are shown in Table \ref{tab:task21}-\ref{tab:task25}.  The prompts for self-refinement generation are shown in Table \ref{tab:task31}-\ref{tab:task35}. Note that we have distinct prompts for two code generation datasets, HumanEval and DS-1000, since a prompt format is already provided for DS-1000 in its repository. For \textit{Auto-J-13B} and \textit{UltraCM-13B} whose input prompt formats are given, we directly adopt their prompt format in data construction, training and testing, as shown in Table \ref{tab:autoj} and \ref{tab:ucm}. For the rest base models, we use the designed input prompt formats.

The prompt for judging preferences for refinements are shown in Table \ref{tab:prompt1}-\ref{tab:prompt5}, and the prompt for directly judging preferences for critiques are shown in Table \ref{tab:prompt11}-\ref{tab:prompt15}.

\section{Evaluation Prompts}
\label{app:evalprompt}

Our evaluation process involves GPT-4 scoring and preference evaluation, and we design distinct prompts for each of the evaluation settings and task. The prompts for preference judgment and preference scoring are shown in Table \ref{tab:prompt1}-\ref{tab:prompt5}, and the prompts for response quality scoring are shown in Table \ref{tab:prompt6}-\ref{tab:prompt10}.

\begin{table*}[ht]
\centering
\begin{tabular}{p{0.95\textwidth}}
\toprule
\textit{Prompts for Initial Response Generation }\\

\textbf{Dialog Generation}\par
\textit{Directly adopt the original prompt}\\

\textbf{Summarization - TL;DR}\par
Summarize the following Reddit Post: \par
SUBREDDIT: \{subreddit\} \par
TITLE: \{title\} \par
POST: \{post\} \par
Your Summary:\\

\textbf{Summarization - CNN-DailyMail}\par
Summarize the following news: \par
\{news\} \par
Your Summary:\\

\textbf{Question Answering - Multiple Choices (Commonsense QA, ARC-Challenge)}\par
Question: \{prompt\}\par
Answer choices: \{choices\}\par
Explain your reasoning. You must choose only one option from A to E. Your final answer should be a single letter from A to E, in the form [[answer]], at the end of your response. Your answer:\\

\textbf{Question Answering - Others (Trivia-QA, ELI5, AmbigQA)}\par
Question: \{prompt\} \par
Your Answer:\\

\textbf{Mathematical Reasoning - Multiple Choices (MathQA, AQuA)}\par
Can you solve the following math problem? \{prompt\}\par
Answer choices: \{choices\}\par
Explain your reasoning. You must choose only one option from A to E. Your final answer should be a single letter from A to E, in the form [[answer]], at the end of your response. Your answer:\\

\textbf{Mathematical Reasoning - Table (TabMWP)}\par
Can you solve the following math problem according to the information given in the table? \{prompt\}\par
Table: \{table\underline{ }title\}\par
\{table\underline{ }content\}\par
Explain your reasoning. Your final answer should be a single numerical number, in the form $\backslash$boxed\{answer\}, at the end of your response.\\

\textbf{Mathematical Reasoning - Others (TheoremQA)}\par
Can you solve the following math problem? \{prompt\}\\

Explain your reasoning. Your final answer should be a single numerical number, in the form $\backslash$boxed\{answer\}, at the end of your response.\\

\textbf{Code Generation - HumanEval}\par
Complete the code following the instruction given in the comment, as well as the format and the indentation.\par
\{prompt\}\\

\textbf{Code Generation - DS-1000}\par
Write a short code to solve the problem following the given format and indentation. Place the executable code between <code> and </code> tags, without any other non-executable things.\par
\{prompt\}\\

\bottomrule
\end{tabular}
\caption{Prompts for collecting initial responses from actor models.}
\label{tab:task0}
\end{table*}

\begin{table*}[ht]
\centering
\begin{tabular}{p{0.95\textwidth}}
\toprule
\textit{Prompts for Critique Generation}\\

\textbf{Dialog Generation}\par
You are an impartial judge. You are shown a dialog between a human and an AI assistant. Comment on the assistant's response to the dialog according to the criteria of helpful, harmless and correct. After that, provide suggestions for improving the assistant's response according to your comment, starting with "Suggestions for improvement:".\\
----------------\\
\{prompt\} \\
----------------\\
Assistant's Response: \{response\}\\
----------------\\
Your comment:\\
\bottomrule
\end{tabular}
\caption{Prompts for collecting critiques of dialog generation tasks, for \textit{LLaMA-2-7B-Chat}, \textit{LLaMA-2-13B-Chat} and \textit{LLaMA-3-8B-Instruct} base and trained models.}
\label{tab:task11}
\end{table*}

\begin{table*}[ht]
\centering
\begin{tabular}{p{0.95\textwidth}}
\toprule
\textbf{Summarization - TL;DR}\par
You are an impartial judge. You are shown a Reddit post and a summary. Comment on the summary by finding if it misses any key information from the post or contains any unnecessary information. After that, provide suggestions for improving the summary according to your comment, starting with "Suggestions for improvement:".\\
----------------\\
SUBREDDIT: \{subreddit\} \par
TITLE: \{title\} \par
POST: \{post\} \\
----------------\\
Assistant's Summary: \{response\}\\
----------------\\
Your comment:\\

\textbf{Summarization - CNN-DailyMail}\par
You are an impartial judge. You are shown a piece of news and a summary. Comment on the summary by finding if it misses any key information from the post or contains any unnecessary information. After that, provide suggestions for improving the summary according to your comment, starting with "Suggestions for improvement:".\\
----------------\\
\{news\} \\
----------------\\
Assistant's Summary: \{response\}\\
----------------\\
Your comment:\\
\bottomrule
\end{tabular}
\caption{Prompts for collecting critiques of summarization tasks, for \textit{LLaMA-2-7B-Chat}, \textit{LLaMA-2-13B-Chat} and \textit{LLaMA-3-8B-Instruct} base and trained models.}
\label{tab:task12}
\end{table*}

\begin{table*}[ht]
\centering
\begin{tabular}{p{0.95\textwidth}}
\toprule
\textbf{Question Answering - Multiple Choices (Commonsense QA, ARC-Challenge)}\par
You are an impartial judge. You are shown a question and an answer. Comment on the answer and find problems with it. After that, provide suggestions for improving the answer according to your comment, starting with "Suggestions for improvement:".\\
----------------\\
Question: \{prompt\}\par
Answer choices: \{choices\}\\
----------------\\
Assistant's Answer: \{response\}\\
----------------\\
Your comment:\\

\textbf{Question Answering - Others (Trivia-QA, ELI5, AmbigQA)}\par
You are an impartial judge. You are shown a question and an answer. Comment on the answer and find problems with it. After that, provide suggestions for improving the answer according to your comment, starting with "Suggestions for improvement:".\\
----------------\\
Question: \{prompt\}\\
----------------\\
Assistant's Answer: \{response\}\\
----------------\\
Your comment:\\
\bottomrule
\end{tabular}
\caption{Prompts for collecting critiques of question answering tasks, for \textit{LLaMA-2-7B-Chat}, \textit{LLaMA-2-13B-Chat} and \textit{LLaMA-3-8B-Instruct} base and trained models.}
\label{tab:task13}
\end{table*}

\begin{table*}[ht]
\centering
\begin{tabular}{p{0.95\textwidth}}
\toprule
\textbf{Mathematical Reasoning - Multiple Choices (MathQA, AQuA)}\par
You are an expert on mathematics. You are shown a math problem and the answer to it. Comment on the answer and find problems with it. After that, provide suggestions for improving the answer according to your comment, starting with "Suggestions for improvement:".\\
----------------\\
Problem: \{prompt\}\par
Answer choices: \{choices\}\\
----------------\\
Assistant's Answer: \{response\}\\
----------------\\
Your comment:\\

\textbf{Mathematical Reasoning - Table (TabMWP)}\par
You are an expert on mathematics. You are shown a math problem and the answer to it. Comment on the answer and find problems with it. After that, provide suggestions for improving the answer according to your comment, starting with "Suggestions for improvement:".\\
----------------\\
Problem: \{prompt\}\par
Table: \{table\underline{ }title\}\par
\{table\underline{ }content\}\\
----------------\\
Assistant's Answer: \{response\}\\
----------------\\
Your comment:\\

\textbf{Mathematical Reasoning - Others (TheoremQA)}\par
You are an expert on mathematics. You are shown a math problem and the answer to it. Comment on the answer and find problems with it. After that, provide suggestions for improving the answer according to your comment, starting with "Suggestions for improvement:".\\
----------------\\
Problem: \{prompt\}\\
----------------\\
Assistant's Answer: \{response\}\\
----------------\\
Your comment:\\
\bottomrule
\end{tabular}
\caption{Prompts for collecting critiques of mathematical reasoning tasks, for \textit{LLaMA-2-7B-Chat}, \textit{LLaMA-2-13B-Chat} and \textit{LLaMA-3-8B-Instruct} base and trained models.}
\label{tab:task14}
\end{table*}

\begin{table*}[ht]
\centering
\begin{tabular}{p{0.95\textwidth}}
\toprule
\textbf{Code Generation - HumanEval}\par
You are an expert on computer programming. You are shown a code completion according to the requirements presented in the comment line of the code. Evaluate the correctness and readability of the code, and find if it meet the presented requirements. After that, provide suggestions for improving the code according to your evaluation, starting with "Suggestions for improvement:".\\
----------------\\
\{prompt\} \{response\}\\
----------------\\
Your evaluation:\\

\textbf{Code Generation - DS-1000}\par
You are an expert on computer programming. You are shown a code that proposes to solve the coding problem. Evaluate the correctness and readability of the code completion, and find if it meet the presented requirements. Remember not to comment on anything between ’A:’ and ’BEGIN SOLUTION’. After that, provide suggestions for improving the code according to your evaluation, starting with "Suggestions for improvement":\\
----------------\\
\{prompt\} \{response\}\\
----------------\\
Your evaluation:\\

\bottomrule
\end{tabular}
\caption{Prompts for collecting critiques of code generation tasks, for \textit{LLaMA-2-7B-Chat}, \textit{LLaMA-2-13B-Chat} and \textit{LLaMA-3-8B-Instruct} base and trained models.}
\label{tab:task15}
\end{table*}

\begin{table*}[ht]
\centering
\begin{tabular}{p{0.95\textwidth}}
\toprule
\textit{Auto-J-13B Prompt}\\

[INST] Write critiques for a submitted response on a given user's query, and grade the response:\par
[BEGIN DATA]\par
***\par
[Query]: \{prompt\}\par
***\par
[Response]: \{answer\}
***\par
[END DATA]\\
Write critiques for this response. After that, you should give a final rating for the response on a scale of 1 to 10 by strictly following this format: "[[rating]]", for example: "Rating: [[5]]". [/INST]\\
\bottomrule

\end{tabular}
\caption{Prompt for collecting critiques for \textit{Auto-J-13B} base and trained models.}
\label{tab:autoj}
\end{table*}

\begin{table*}[ht]
\centering
\begin{tabular}{p{0.95\textwidth}}
\toprule
\textit{UltraCM-13B Prompt}\\

Given my answer to an instruction, your role is to provide specific and constructive feedback for me. You should find the best way for me to learn from your feedback and improve my performance. \\

You should consider multiple aspects of my answer, including helpfulness, truthfulness, honesty, and to what extent the answer follows instructions.\par
---\\
\#\#\# Instruction\par
\{prompt\}\\

\#\#\# Answer\par
\{answer\}\par
---\\
Please act as a teacher and provide specific and constructive feedback. Besides describing the weaknesses of the answer, you should also provide specific suggestions to guide me toward understanding how to improve. Please note, however, that your suggestions should help me better complete the instructions, but you should not introduce new requirements that are not mentioned in the instructions. Your feedback should focus on enhancing my ability to think critically and respond accurately. However, never explicitly provide the reference answer, nor do polite phrases be required. Only respond with concise feedback in chat style. Finally, score the overall quality of the answer from 1 to 10, where 1 is the worst and 10 is the best.\\
*Format*\par
\#\#\# Feedback\par
Overall Score: [1-10]\par
[Your feedback]\\
---\\
\#\#\# Feedback\par
Overall Score:\\
\bottomrule

\end{tabular}
\caption{Prompts for collecting critiques for \textit{Ultra-CM-13B} base and trained models.}
\label{tab:ucm}
\end{table*}

\begin{table*}[ht]
\centering
\begin{tabular}{p{0.95\textwidth}}
\toprule
\textit{Prompts for Refined Response Generation}\\

\textbf{Dialog Generation}\par
You are shown a dialog between a human and an AI assistant. An impartial judge on AI assistants has made comments on the assistant's response to the dialog. Please revise the assistant's response to improve its quality according to the suggestions for improvement provided in the comment, starting with "My revised response:".\\
----------------\\
\{prompt\} \\
----------------\\
Assistant's Response: \{response\}\\
----------------\\
Comment by the judge: \{critique\}\\
----------------\\
Your revision:\\
\bottomrule
\end{tabular}
\caption{Prompts for collecting refined responses of dialog generation tasks.}
\label{tab:task21}
\end{table*}

\begin{table*}[ht]
\centering
\begin{tabular}{p{0.95\textwidth}}
\toprule
\textbf{Summarization - TL;DR}\par
You are shown a Reddit post and a summary. An impartial judge has made comments on the summary. Please revise the summary to improve its quality according to the suggestions for improvement provided in the comment, starting with "My revised summary:".\\
----------------\\
SUBREDDIT: \{subreddit\} \par
TITLE: \{title\} \par
POST: \{post\} \\
----------------\\
Original Summary: \{response\}\\
----------------\\
Comment by the judge: \{critique\}\\
----------------\\
Your revision:\\

\textbf{Summarization - CNN-DailyMail}\par
You are shown a piece of news and a summary. An impartial judge has made comments on the summary. Please revise the summary to improve its quality according to the suggestions for improvement provided in the comment, starting with "My revised summary:".\\
----------------\\
\{news\} \\
----------------\\
Original Summary: \{response\}\\
----------------\\
Comment by the judge: \{critique\}\\
----------------\\
Your revision:\\
\bottomrule
\end{tabular}
\caption{Prompts for collecting refined responses of summarization tasks.}
\label{tab:task22}
\end{table*}

\begin{table*}[ht]
\centering
\begin{tabular}{p{0.95\textwidth}}
\toprule
\textbf{Question Answering - Multiple Choices (Commonsense QA, ARC-Challenge)}\par
You are shown a question and an answer. An impartial judge has made comments on the answer. Please revise the answer to improve its quality according to the suggestions for improvement provided in the comment, starting with "My revised answer:". \\
----------------\\
Question: \{prompt\}\par
Answer choices: \{choices\}\\
----------------\\
Original Answer: \{response\}\\
----------------\\
Comment by the judge: \{critique\}\\
----------------\\
Your revision:\\

\textbf{Question Answering - Others (Trivia-QA, ELI5, AmbigQA)}\par
You are shown a question and an answer. An impartial judge has made comments on the answer. Please revise the answer to improve its quality according to the suggestions for improvement provided in the comment, starting with "My revised answer:". \\
----------------\\
Question: \{prompt\}\\
----------------\\
Original Answer: \{response\}\\
----------------\\
Comment by the judge: \{critique\}\\
----------------\\
Your revision:\\
\bottomrule
\end{tabular}
\caption{Prompts for collecting refined responses of question answering tasks.}
\label{tab:task23}
\end{table*}

\begin{table*}[ht]
\centering
\begin{tabular}{p{0.95\textwidth}}
\toprule
\textbf{Mathematical Reasoning - Multiple Choices (MathQA, AQuA)}\par
You are shown a math problem and an answer. An expert on mathematics has made comments on the answer. Please revise the answer to improve its quality according to the suggestions for improvement provided in the comment, starting with "My revised answer:".\\
----------------\\
Problem: \{prompt\}\par
Answer choices: \{choices\}\\
----------------\\
Original Answer: \{response\}\\
----------------\\
Comment by the expert: \{critique\}\\
----------------\\
Your revision:\\

\textbf{Mathematical Reasoning - Table (TabMWP)}\par
You are shown a math problem and an answer. An expert on mathematics has made comments on the answer. Please revise the answer to improve its quality according to the suggestions for improvement provided in the comment, starting with "My revised answer:".\\
----------------\\
Problem: \{prompt\}\par
Table: \{table\underline{ }title\}\par
\{table\underline{ }content\}\\
----------------\\
Original Answer: \{response\}\\
----------------\\
Comment by the expert: \{critique\}\\
----------------\\
Your revision:\\

\textbf{Mathematical Reasoning - Others (TheoremQA)}\par
You are shown a math problem and an answer. An expert on mathematics has made comments on the answer. Please revise the answer to improve its quality according to the suggestions for improvement provided in the comment, starting with "My revised answer:".\\
----------------\\
Problem: \{prompt\}\\
----------------\\
Original Answer: \{response\}\\
----------------\\
Comment by the expert: \{critique\}\\
----------------\\
Your revision:\\
\bottomrule
\end{tabular}
\caption{Prompts for collecting refined responses of mathematical reasoning tasks.}
\label{tab:task24}
\end{table*}

\begin{table*}[ht]
\centering
\begin{tabular}{p{0.95\textwidth}}
\toprule
\textbf{Code Generation - HumanEval}\par
You are shown a code completion according to the requirements presented in the comment. An expert on computer programming has made critiques and advice for improvement on the code. Please revise the code completion to improve its quality according to the suggestions for improvement provided in the critique, starting with "My revised code:". \\
---------------Original Code---------------\\
\{prompt\} \{response\}\\
---------------Critiques and Advice--------\\
\{critique\}\\
---------------Your Revision---------------\\
\{prompt\}\\

\textbf{Code Generation - DS-1000}\par
You are shown a code that proposes to solve the coding problem. An expert on computer programming has made critiques and advice for improvement on the code. Please revise the code completion to improve its quality according to the suggestions for improvement provided in the critique, starting with "My revised code:". \\
---------------Original Code---------------\\
\{prompt\} \{response\}\\
---------------Critiques and Advice--------\\
\{critique\}\\
---------------Your Revision---------------\\
\\

\bottomrule
\end{tabular}
\caption{Prompts for collecting refined responses of code generation tasks.}
\label{tab:task25}
\end{table*}

\begin{table*}[ht]
\centering
\begin{tabular}{p{0.95\textwidth}}
\toprule
\textit{Prompts for Self-Refinement Response Generation}\\

\textbf{Dialog Generation}\par
You are shown a dialog between a human and an AI assistant. Please revise the assistant's response to improve its quality according to your analysis, starting with "My revised response:". \\
----------------\\
\{prompt\} \\
----------------\\
Assistant's Response: \{response\}\\
----------------\\
Your revision:\\
\bottomrule
\end{tabular}
\caption{Prompts for collecting self-refinement responses of dialog generation tasks.}
\label{tab:task31}
\end{table*}

\begin{table*}[ht]
\centering
\begin{tabular}{p{0.95\textwidth}}
\toprule
\textbf{Summarization - TL;DR}\par
You are shown a Reddit post and a summary of it. Please revise the summary to improve its quality according to your analysis, starting with "My revised summary:". \\
----------------\\
SUBREDDIT: \{subreddit\} \par
TITLE: \{title\} \par
POST: \{post\} \\
----------------\\
Original Summary: \{response\}\\
----------------\\
Your revision:\\

\textbf{Summarization - CNN-DailyMail}\par
You are shown a piece of news and a summary of it. Please revise the summary to improve its quality according to your analysis, starting with "My revised summary:".\\
----------------\\
\{news\} \\
----------------\\
Original Summary: \{response\}\\
----------------\\
Your revision:\\
\bottomrule
\end{tabular}
\caption{Prompts for collecting self-refinement responses of summarization tasks.}
\label{tab:task32}
\end{table*}

\begin{table*}[ht]
\centering
\begin{tabular}{p{0.95\textwidth}}
\toprule
\textbf{Question Answering - Multiple Choices (Commonsense QA, ARC-Challenge)}\par
You are shown a question and an answer. Please revise the answer to improve its quality according to your analysis, starting with "My revised answer:". \\
----------------\\
Question: \{prompt\}\par
Answer choices: \{choices\}\\
----------------\\
Original Answer: \{response\}\\
----------------\\
Your revision:\\

\textbf{Question Answering - Others (Trivia-QA, ELI5, AmbigQA)}\par
You are shown a question and an answer. Please revise the answer to improve its quality according to your analysis, starting with "My revised answer:".\\
----------------\\
Question: \{prompt\}\\
----------------\\
Original Answer: \{response\}\\
----------------\\
Your revision:\\
\bottomrule
\end{tabular}
\caption{Prompts for collecting self-refinement responses of question answering tasks.}
\label{tab:task33}
\end{table*}

\begin{table*}[ht]
\centering
\begin{tabular}{p{0.95\textwidth}}
\toprule
\textbf{Mathematical Reasoning - Multiple Choices (MathQA, AQuA)}\par
You are shown a math problem and an answer. Please revise the answer to improve its quality according to your analysis, starting with "My revised answer:".\\
----------------\\
Problem: \{prompt\}\par
Answer choices: \{choices\}\\
----------------\\
Original Answer: \{response\}\\
----------------\\
Your revision:\\

\textbf{Mathematical Reasoning - Table (TabMWP)}\par
You are shown a math problem and an answer. Please revise the answer to improve its quality according to your analysis, starting with "My revised answer:".\\
----------------\\
Problem: \{prompt\}\par
Table: \{table\underline{ }title\}\par
\{table\underline{ }content\}\\
----------------\\
Original Answer: \{response\}\\
----------------\\
Your revision:\\

\textbf{Mathematical Reasoning - Others (TheoremQA)}\par
You are shown a math problem and an answer. Please revise the answer to improve its quality according to your analysis, starting with "My revised answer:".\\
----------------\\
Problem: \{prompt\}\\
----------------\\
Original Answer: \{response\}\\
----------------\\
Your revision:\\
\bottomrule
\end{tabular}
\caption{Prompts for collecting self-refinement responses of mathematical reasoning tasks.}
\label{tab:task34}
\end{table*}

\begin{table*}[ht]
\centering
\begin{tabular}{p{0.95\textwidth}}
\toprule
\textbf{Code Generation - HumanEval}\par
You are shown a code completion according to the requirements presented in the comment. Please revise the code to make it more correct and readable, starting with "My revised code:".\\
---------------Original Code---------------\\
\{prompt\} \{response\}\\
---------------Your Revision---------------\\
\{prompt\}\\

\textbf{Code Generation - DS-1000}\par
You are shown a code that proposes to solve the coding problem. Please revise the code to make it more correct and readable, starting with "My revised code:". \\
---------------Original Code---------------\\
\{prompt\} \{response\}\\
---------------Your Revision---------------\\
\\

\bottomrule
\end{tabular}
\caption{Prompts for collecting self-refinement responses of code generation tasks.}
\label{tab:task35}
\end{table*}

\begin{table*}[ht]
\centering
\begin{tabular}{p{0.95\textwidth}}
\toprule
\textit{Dialog Generation}

\textbf{[SYSTEM]}\par
Please act as an impartial judge and evaluate the quality of the responses provided by two AI assistants to the conversation displayed below. You should choose the assistant that follows the user's instructions better. Your evaluation should consider factors such as the helpfulness, relevance, accuracy, depth, creativity, and level of detail of their responses. You should focus on who provides a better response. Begin your evaluation by comparing the responses of the two assistants and provide a short explanation. Avoid any position biases and ensure that the order in which the responses were presented does not influence your decision. Do not allow the length of the responses to influence your evaluation. Do not favor certain names of the assistants. Be as objective as possible. After providing your explanation, output your final verdict by strictly following this format: "[[A]]" if assistant A is better, "[[B]]" if assistant B is better, and "[[C]]" for a tie.\\

\textbf{[USER]}\par
[Conversation]\par
\{prompt\} \\

[The Start of Assistant A's Response]\par
\{answer\underline{ }0\}\par
[The End of Assistant A's Response] \\

[The Start of Assistant B's Response]\par
\{answer\underline{ }1\}\par
[The End of Assistant B's Response]\\
\bottomrule

\end{tabular}
\caption{Prompt for refinement preference judgment and evaluation for dialog generation tasks.}
\label{tab:prompt1}
\end{table*}

\begin{table*}[ht]
\centering
\begin{tabular}{p{0.95\textwidth}}
\toprule
\textit{Summarization}

\textbf{[SYSTEM]}\par
Please act as an impartial judge and evaluate the quality of the summaries provided by two AI assistants to the \{kind\} displayed below. Your evaluation should consider whether their summaries include all key information from the original article and avoid false or unnecessary sentences. Your should decide which assistant's summary is better. Begin your evaluation by comparing both assistants' summaries and provide a short explanation. Avoid any position biases and ensure that the order in which the responses were presented does not influence your decision. Do not allow the length of the responses to influence your evaluation. Do not favor certain names of the assistants. Be as objective as possible. After providing your explanation, output your final verdict by strictly following this format: "[[A]]" if assistant A is better, "[[B]]" if assistant B is better, and "[[C]]" for a tie.\\

\textbf{[USER]}\par
[\{kind\}]\par
\{prompt\} \\

[The Start of Assistant A's Summary]\par
\{answer\underline{ }0\}\par
[The End of Assistant A's Summary] \\

[The Start of Assistant B's Summary]\par
\{answer\underline{ }1\}\par
[The End of Assistant B's Summary]\\
\bottomrule

\end{tabular}
\caption{Prompt for refinement preference judgment and evaluation for summarization tasks. The ``kind'' field will be ``Reddit post'' or ``News'', conditioned to whether this prompt is from TL;DR or CNN-DailyMail dataset.}
\label{tab:prompt2}
\end{table*}

\begin{table*}[ht]
\centering
\begin{tabular}{p{0.95\textwidth}}
\toprule
\textit{Question Answering}

\textbf{[SYSTEM]}\par
Please act as an impartial judge and evaluate the quality of the answers provided by two AI assistants to the user question displayed below. You should choose the assistant that follows the user's instructions and answers the user's question better. Your evaluation should consider factors such as the helpfulness, relevance, accuracy, depth, creativity, and level of detail of their responses. Begin your evaluation by comparing the two responses and provide a short explanation. Avoid any position biases and ensure that the order in which the responses were presented does not influence your decision. Do not allow the length of the responses to influence your evaluation. Do not favor certain names of the assistants. Be as objective as possible. After providing your explanation, output your final verdict by strictly following this format: "[[A]]" if assistant A is better, "[[B]]" if assistant B is better, and "[[C]]" for a tie.\\

\textbf{[USER]}\par
[User Question]\par
\{prompt\} \\

[The Start of Assistant A's Answer]\par
\{answer\underline{ }0\}\par
[The End of Assistant A's Answer] \\

[The Start of Assistant B's Answer]\par
\{answer\underline{ }1\}\par
[The End of Assistant B's Answer]\\
\bottomrule

\end{tabular}
\caption{Prompt for refinement preference judgment and evaluation for question answering tasks.}
\label{tab:prompt3}
\end{table*}

\begin{table*}[ht]
\centering
\begin{tabular}{p{0.95\textwidth}}
\toprule
\textit{Mathematical Reasoning}

\textbf{[SYSTEM]}\par
Please act as an impartial judge and evaluate the quality of the answers provided by two AI assistants to the math problem displayed below. Your evaluation should consider correctness and helpfulness. You will be given a reference answer, assistant A's answer, and assistant B's answer. Your job is to evaluate which assistant's answer is better. Begin your evaluation by comparing both assistants' answers with the reference answer. Identify and correct any mistakes. Avoid any position biases and ensure that the order in which the responses were presented does not influence your decision. Do not allow the length of the responses to influence your evaluation. Do not favor certain names of the assistants. Be as objective as possible. After providing your explanation, output your final verdict by strictly following this format: "[[A]]" if assistant A is better, "[[B]]" if assistant B is better, and "[[C]]" for a tie.\\

\textbf{[USER]}\par
[Math Problem]\par
\{prompt\} \\

[The Start of Reference Answer]\par
\{ref\underline{ }answer\}\par
[The End of Reference Answer] \\

[The Start of Assistant A's Answer]\par
\{answer\underline{ }0\}\par
[The End of Assistant A's Answer] \\

[The Start of Assistant B's Answer]\par
\{answer\underline{ }1\}\par
[The End of Assistant B's Answer]\\
\bottomrule

\end{tabular}
\caption{Prompt for refinement preference judgment and evaluation for mathematical reasoning tasks.}
\label{tab:prompt4}
\end{table*}

\begin{table*}[ht]
\centering
\begin{tabular}{p{0.95\textwidth}}
\toprule
\textit{Code Generation}

\textbf{[SYSTEM]}\par
Please act as an impartial judge and evaluate the quality of the code provided by two AI assistants to the requirements displayed below. Your evaluation should consider correctness and helpfulness. Your should decide which assistant's provided code is better. Begin your evaluation by comparing both assistants' codes and provide a short explanation. Identify and correct any mistakes. Avoid any position biases and ensure that the order in which the responses were presented does not influence your decision. Do not allow the length of the responses to influence your evaluation. Do not favor certain names of the assistants. Be as objective as possible. After providing your explanation, output your final verdict by strictly following this format: "[[A]]" if assistant A is better, "[[B]]" if assistant B is better, and "[[C]]" for a tie.\\

\textbf{[USER]}\par
[Code Requirements]\par
\{prompt\} \\

[The Start of Assistant A's Code]\par
\{answer\underline{ }0\}\par
[The End of Assistant A's Code] \\

[The Start of Assistant B's Code]\par
\{answer\underline{ }1\}\par
[The End of Assistant B's Code]\\
\bottomrule

\end{tabular}
\caption{Prompt for refinement preference judgment and evaluation for code generation tasks.}
\label{tab:prompt5}
\end{table*}

\begin{table*}[ht]
\centering
\begin{tabular}{p{0.95\textwidth}}
\toprule
\textit{Dialog Generation}

Please act as an impartial judge and evaluate the quality of the critiques provided by two AI assistants for my response to the conversation displayed below. Your evaluation should focus on the quality, clarity, and constructiveness of the critiques, particularly the "Suggestions for improvement" field. \\
You will be given the conversation, my response, assistant A's critique, and assistant B's critique. Your job is to assess which assistant's critique is better based on the following criteria: \\
1. **Accuracy:** Does the critique accurately identify any issues with my response? Are any mistakes or shortcomings in my response correctly pointed out? \par
2. **Clarity:** Is the critique clearly written, easy to understand, and well-structured? Does it explain the issues in a way that is accessible and straightforward? \par
3. **Constructiveness:** Does the critique provide practical and actionable suggestions for improvement? Are the suggestions detailed, specific, and relevant to the issues identified? \par
4. **Objectivity:** Is the critique unbiased and impartial? Does it focus solely on the quality of my response and avoid unnecessary personal opinions or judgments? \par
5. **Thoroughness:** Does the critique cover all significant aspects of my response, or does it overlook any important issues? Does it delve into the reasoning behind the suggestions for improvement? \par
6. **Tone:** Is the critique delivered in a respectful and professional tone, avoiding any condescension or harshness? \\
You should focus particularly on the "Suggestions for improvement" field in each critique and evaluate how well each assistant has provided guidance to improve the response. Avoid being influenced by the length of the critiques or the order in which they are presented. Do not favor one assistant over the other based on irrelevant factors. Be as objective as possible. After providing your explanation, output your final verdict by strictly following this format: "[[A]]" if assistant A is better, "[[B]]" if assistant B is better, and "[[C]]" for a tie.\\

\textbf{[USER]}\par
[Conversation]\par
\{prompt\} \\

[The Start of My Response]\par
\{answer\}\par
[The End of My Response] \\

[The Start of Assistant A's Critique]\par
\{critique\underline{ }0\}\par
[The End of Assistant A's Critique] \\

[The Start of Assistant B's Critique]\par
\{critique\underline{ }1\}\par
[The End of Assistant B's Critique]\\
\bottomrule

\end{tabular}
\caption{Prompt for critique preference judgment for dialog generation tasks.}
\label{tab:prompt11}
\end{table*}

\begin{table*}[ht]
\centering
\begin{tabular}{p{0.95\textwidth}}
\toprule
\textit{Summarization}

\textbf{[SYSTEM]}\par
Please act as an impartial judge and evaluate the quality of the critiques provided by two AI assistants for my summary to the \{kind\} displayed below. Your evaluation should focus on the quality, clarity, and constructiveness of the critiques, particularly the "Suggestions for improvement" field. \\
You will be given the \{kind\}, my summary, assistant A's critique, and assistant B's critique. Your job is to assess which assistant's critique is better based on the following criteria: \\
1. **Accuracy:** Does the critique accurately identify any issues with my summary? Are any mistakes or shortcomings in my summary correctly pointed out? \par
2. **Clarity:** Is the critique clearly written, easy to understand, and well-structured? Does it explain the issues in a way that is accessible and straightforward? \par
3. **Constructiveness:** Does the critique provide practical and actionable suggestions for improvement? Are the suggestions detailed, specific, and relevant to the issues identified? \par
4. **Objectivity:** Is the critique unbiased and impartial? Does it focus solely on the quality of my summary and avoid unnecessary personal opinions or judgments? \par
5. **Thoroughness:** Does the critique cover all significant aspects of my summary, or does it overlook any important issues? Does it delve into the reasoning behind the suggestions for improvement? \par
6. **Tone:** Is the critique delivered in a respectful and professional tone, avoiding any condescension or harshness? \\
You should focus particularly on the "Suggestions for improvement" field in each critique and evaluate how well each assistant has provided guidance to improve the summary. Avoid being influenced by the length of the critiques or the order in which they are presented. Do not favor one assistant over the other based on irrelevant factors. Be as objective as possible. After providing your explanation, output your final verdict by strictly following this format: "[[A]]" if assistant A is better, "[[B]]" if assistant B is better, and "[[C]]" for a tie.\\\\

\textbf{[USER]}\par
[\{kind\}]\par
\{prompt\} \\

[The Start of My Summary]\par
\{answer\}\par
[The End of My Summary] \\

[The Start of Assistant A's Critique]\par
\{critique\underline{ }0\}\par
[The End of Assistant A's Critique] \\

[The Start of Assistant B's Critique]\par
\{critique\underline{ }1\}\par
[The End of Assistant B's Critique]\\
\bottomrule

\end{tabular}
\caption{Prompt for critique preference judgment for summarization tasks. The ``kind'' field will be ``Reddit post'' or ``News'', conditioned to whether this prompt is from TL;DR or CNN-DailyMail dataset.}
\label{tab:prompt12}
\end{table*}

\begin{table*}[ht]
\centering
\begin{tabular}{p{0.95\textwidth}}
\toprule
\textit{Question Answering}

\textbf{[SYSTEM]}\par
Please act as an impartial judge and evaluate the quality of the critiques provided by two AI assistants for my answer to the question displayed below. Your evaluation should focus on the quality, clarity, and constructiveness of the critiques, particularly the "Suggestions for improvement" field. \\
You will be given the question, my answer, assistant A's critique, and assistant B's critique. Your job is to assess which assistant's critique is better based on the following criteria: \\
1. **Accuracy:** Does the critique accurately identify any issues with my answer? Are any mistakes or shortcomings in my answer correctly pointed out? \par
2. **Clarity:** Is the critique clearly written, easy to understand, and well-structured? Does it explain the issues in a way that is accessible and straightforward? \par
3. **Constructiveness:** Does the critique provide practical and actionable suggestions for improvement? Are the suggestions detailed, specific, and relevant to the issues identified? \par
4. **Objectivity:** Is the critique unbiased and impartial? Does it focus solely on the quality of my answer and avoid unnecessary personal opinions or judgments? \par
5. **Thoroughness:** Does the critique cover all significant aspects of my answer, or does it overlook any important issues? Does it delve into the reasoning behind the suggestions for improvement? \par
6. **Tone:** Is the critique delivered in a respectful and professional tone, avoiding any condescension or harshness? \\
You should focus particularly on the "Suggestions for improvement" field in each critique and evaluate how well each assistant has provided guidance to improve the answer. Avoid being influenced by the length of the critiques or the order in which they are presented. Do not favor one assistant over the other based on irrelevant factors. Be as objective as possible. After providing your explanation, output your final verdict by strictly following this format: "[[A]]" if assistant A is better, "[[B]]" if assistant B is better, and "[[C]]" for a tie.\\

\textbf{[USER]}\par
[User Question]\par
\{prompt\} \\

[The Start of My Answer]\par
\{answer\}\par
[The End of My Answer] \\

[The Start of Assistant A's Critique]\par
\{critique\underline{ }0\}\par
[The End of Assistant A's Critique] \\

[The Start of Assistant B's Critique]\par
\{critique\underline{ }1\}\par
[The End of Assistant B's Critique]\\
\bottomrule

\end{tabular}
\caption{Prompt for critique preference judgment for question answering tasks.}
\label{tab:prompt13}
\end{table*}

\begin{table*}[ht]
\centering
\begin{tabular}{p{0.95\textwidth}}
\toprule
\textit{Mathematical Reasoning}

\textbf{[SYSTEM]}\par
"Please act as an impartial judge and evaluate the quality of the critiques provided by two AI assistants for my answer to the math problem displayed below. Your evaluation should focus on the quality, clarity, and constructiveness of the critiques, particularly the "Suggestions for improvement" field. \\
You will be given the question, my answer, the reference answer, assistant A's critique, and assistant B's critique. Your job is to assess which assistant's critique is better based on the following criteria:\\
1. **Accuracy:** Does the critique accurately identify any issues with my answer? Are any mistakes or shortcomings in my answer correctly pointed out? \par
2. **Clarity:** Is the critique clearly written, easy to understand, and well-structured? Does it explain the issues in a way that is accessible and straightforward? \par
3. **Constructiveness:** Does the critique provide practical and actionable suggestions for improvement? Are the suggestions detailed, specific, and relevant to the issues identified? \par
4. **Objectivity:** Is the critique unbiased and impartial? Does it focus solely on the quality of my answer and avoid unnecessary personal opinions or judgments? \par
5. **Thoroughness:** Does the critique cover all significant aspects of my answer, or does it overlook any important issues? Does it delve into the reasoning behind the suggestions for improvement? \par
6. **Tone:** Is the critique delivered in a respectful and professional tone, avoiding any condescension or harshness? \\
You should focus particularly on the "Suggestions for improvement" field in each critique and evaluate how well each assistant has provided guidance to improve the answer. Avoid being influenced by the length of the critiques or the order in which they are presented. Do not favor one assistant over the other based on irrelevant factors. Be as objective as possible. After providing your explanation, output your final verdict by strictly following this format: "[[A]]" if assistant A is better, "[[B]]" if assistant B is better, and "[[C]]" for a tie.\\

\textbf{[USER]}\par
[Math Problem]\par
\{prompt\} \\

[The Start of Reference Answer]\par
\{ref\underline{ }answer\}\par
[The End of Reference Answer] \\

[The Start of My Answer]\par
\{answer\}\par
[The End of My Answer] \\

[The Start of Assistant A's Critique]\par
\{critique\underline{ }0\}\par
[The End of Assistant A's Critique] \\

[The Start of Assistant B's Critique]\par
\{critique\underline{ }1\}\par
[The End of Assistant B's Critique]\\
\bottomrule

\end{tabular}
\caption{Prompt for critique preference judgment for mathematical reasoning tasks.}
\label{tab:prompt14}
\end{table*}

\begin{table*}[ht]
\centering
\begin{tabular}{p{0.95\textwidth}}
\toprule
\textit{Code Generation}

\textbf{[SYSTEM]}\par
Please act as an impartial judge and evaluate the quality of the critiques provided by two AI assistants for my code to the requirements displayed below. Your evaluation should focus on the quality, clarity, and constructiveness of the critiques, particularly the "Suggestions for improvement" field. \\
You will be given the question, my code, assistant A's critique, and assistant B's critique. Your job is to assess which assistant's critique is better based on the following criteria:\\
1. **Accuracy:** Does the critique accurately identify any issues with my code? Are any mistakes or shortcomings in my code correctly pointed out? \par
2. **Clarity:** Is the critique clearly written, easy to understand, and well-structured? Does it explain the issues in a way that is accessible and straightforward? \par
3. **Constructiveness:** Does the critique provide practical and actionable suggestions for improvement? Are the suggestions detailed, specific, and relevant to the issues identified? \par
4. **Objectivity:** Is the critique unbiased and impartial? Does it focus solely on the quality of my code and avoid unnecessary personal opinions or judgments? \par
5. **Thoroughness:** Does the critique cover all significant aspects of my code, or does it overlook any important issues? Does it delve into the reasoning behind the suggestions for improvement? \par
6. **Tone:** Is the critique delivered in a respectful and professional tone, avoiding any condescension or harshness? \\
You should focus particularly on the "Suggestions for improvement" field in each critique and evaluate how well each assistant has provided guidance to improve the code. Avoid being influenced by the length of the critiques or the order in which they are presented. Do not favor one assistant over the other based on irrelevant factors. Be as objective as possible. After providing your explanation, output your final verdict by strictly following this format: "[[A]]" if assistant A is better, "[[B]]" if assistant B is better, and "[[C]]" for a tie.\\

\textbf{[USER]}\par
[Code Requirements]\par
\{prompt\} \\

[The Start of My Code]\par
\{answer\}\par
[The End of My Code] \\

[The Start of Assistant A's Critique]\par
\{critique\underline{ }0\}\par
[The End of Assistant A's Critique] \\

[The Start of Assistant B's Critique]\par
\{critique\underline{ }1\}\par
[The End of Assistant B's Critique]\\
\bottomrule

\end{tabular}
\caption{Prompt for critique preference judgment for code generation tasks.}
\label{tab:prompt15}
\end{table*}

\begin{table*}[ht]
\centering
\begin{tabular}{p{0.95\textwidth}}
\toprule
\textit{Response Quality Scoring: Dialog Generation}

\textbf{[SYSTEM]}\par
Please act as an impartial judge and evaluate the quality of the response provided by an AI assistant to the conversation displayed below. Your evaluation should consider factors such as the helpfulness, relevance, accuracy, depth, creativity, and level of detail of the response. Begin your evaluation by providing a short explanation. Do not allow the length of the response to influence your evaluation. Be as objective as possible. After providing your explanation, you must rate the response on a scale of 1 to 10 by strictly following this format: "[[rating]]", for example: "Rating: [[5]]".\\

\textbf{[USER]}\par
[Conversation]\par
\{prompt\} \\

[The Start of Assistant's Response]\par
\{answer\}\par
[The End of Assistant's Response]\\
\bottomrule

\end{tabular}
\caption{Prompt for refinement quality scoring for dialog generation tasks.}
\label{tab:prompt6}
\end{table*}

\begin{table*}[ht]
\centering
\begin{tabular}{p{0.95\textwidth}}
\toprule
\textit{Response Quality Scoring: Summarization}

\textbf{[SYSTEM]}\par
Please act as an impartial judge and evaluate the quality of the summary provided by an AI assistant to the {kind} displayed below. Your evaluation should consider whether the summary include all key information from the original article and avoid false or unnecessary sentences. Begin your evaluation by providing a short explanation. Do not allow the length of the summary to influence your evaluation. Be as objective as possible. After providing your explanation, you must rate the summary on a scale of 1 to 10 by strictly following this format: "[[rating]]", for example: "Rating: [[5]]".\\

\textbf{[USER]}\par
[\{kind\}]\par
\{prompt\} \\

[The Start of Assistant's Summary]\par
\{answer\}\par
[The End of Assistant's Summary] \\
\bottomrule

\end{tabular}
\caption{Prompt for refinement quality scoring for summarization tasks. The ``kind'' field will be ``Reddit post'' or ``News'', conditioned to whether this prompt is from TL;DR or CNN-DailyMail dataset.}
\label{tab:prompt7}
\end{table*}

\begin{table*}[ht]
\centering
\begin{tabular}{p{0.95\textwidth}}
\toprule
\textit{Response Quality Scoring: Question Answering}

\textbf{[SYSTEM]}\par
Please act as an impartial judge and evaluate the quality of the answer provided by an AI assistant to the user question. Your evaluation should consider factors such as the helpfulness, relevance, accuracy, depth, creativity, and level of detail in the answer. Begin your evaluation by providing a short explanation. Do not allow the length of the answer to influence your evaluation. Be as objective as possible. After providing your explanation, you must rate the answer on a scale of 1 to 10 by strictly following this format: "[[rating]]", for example: "Rating: [[5]]".\\

\textbf{[USER]}\par
[User Question]\par
\{prompt\} \\

[The Start of Assistant's Answer]\par
\{answer\}\par
[The End of Assistant's Answer] \\
\bottomrule

\end{tabular}
\caption{Prompt for refinement quality scoring for question answering tasks.}
\label{tab:prompt8}
\end{table*}

\begin{table*}[ht]
\centering
\begin{tabular}{p{0.95\textwidth}}
\toprule
\textit{Response Quality Scoring: Mathematical Reasoning}

\textbf{[SYSTEM]}\par
Please act as an impartial judge and evaluate the quality of the answer provided by an AI assistant to the math problem. Your evaluation should consider correctness and helpfulness. You will be given a reference answer and the assistant's answer. Begin your evaluation by comparing the assistant's answer with the reference answer. Identify and correct any mistakes. Do not allow the length of the answer to influence your evaluation. Be as objective as possible. After providing your explanation, you must rate the answer on a scale of 1 to 10 by strictly following this format: "[[rating]]", for example: "Rating: [[5]]".\\

\textbf{[USER]}\par
[Math Problem]\par
\{prompt\} \\

[The Start of Reference Answer]\par
\{ref\underline{ }answer\}\par
[The End of Reference Answer] \\

[The Start of Assistant's Answer]\par
\{answer\}\par
[The End of Assistant's Answer] \\
\bottomrule

\end{tabular}
\caption{Prompt for refinement quality scoring for mathematical reasoning tasks.}
\label{tab:prompt9}
\end{table*}

\begin{table*}[ht]
\centering
\begin{tabular}{p{0.95\textwidth}}
\toprule
\textit{Response Quality Scoring: Code Generation}

\textbf{[SYSTEM]}\par
Please act as an impartial judge and evaluate the quality of the code provided by an AI assistant to the requirements displayed below. Your evaluation should consider correctness and helpfulness. Begin your evaluation by providing a short explanation. Identify and correct any mistakes. Do not allow the length of the response to influence your evaluation. Be as objective as possible. After providing your explanation, you must rate the code on a scale of 1 to 10 by strictly following this format: "[[rating]]", for example: "Rating: [[5]]".\\

\textbf{[USER]}\par
[Code Requirements]\par
\{prompt\} \\

[The Start of Assistant's Code]\par
\{answer\}\par
[The End of Assistant's Code]\\
\bottomrule

\end{tabular}
\caption{Prompt for refinement quality scoring for code generation tasks.}
\label{tab:prompt10}
\end{table*}

\section{Training Details}
\label{app:training}

In our experiment, we train RCO and DPCO on 5 base critic models: \textit{LLaMA-2-7B-Chat}, \textit{LLaMA-2-13B-Chat}, \textit{LLaMA-3-8B-Instruct}, \textit{Auto-J-13B}, and \textit{UltraCM-13B}. We train the smallest model \textit{LLaMA-2-7B-Chat} on 4 NVIDIA H800 80GB GPUs with a batch size of 2, a gradient accumulation of 4 and 100 warmup steps. 
We train the medium-sized model \textit{LLaMA-3-8B-Instruct} on 4 NVIDIA H800 80GB GPUs with a batch size of 1, a gradient accumulation of 8 and 100 warmup steps. 
For the largest base models \textit{LLaMA-3-8B-Instruct}, \textit{Auto-J-13B}, and \textit{UltraCM-13B}, we train each of these models on 6 NVIDIA H800 80GB GPUs with a batch size of 1, a gradient accumulation of 8 and 50 warmup steps.
Each of these models are fully trained 5 epochs for RCO and 1 epoch for DPCO, ensuring equalized numbers of optimization steps, with a learning rate of $1\times 10^{-6}$ and a linear warmup schedule. 
We use $\beta=0.1$ throughout our study.
For DPCO,  we incorporate all 10,000 prompts to train DPCO (which is $M=5$ times of RCO), in order to ensure equalized preference judgment times for RCO and DPCO. This is because RCO requires $N\times M\times 2$ refinement judgments and DPCO requires $N\times 2$ critique judgments. After filtering out invalid and inconsistent judgments, we gather 56,535 preference pairs for each critic model. These preference pairs are used to train critic models via the DPO algorithm.

We report the training time of RCO and DPCO in Table \ref{tab:time}.

\section{Human Evaluation Details}
\label{app:human}

For the human evaluation, we sample 200 responses from the benchmark, ensuring that each task has 40 responses. 
Given that the code generation task in CriticEval overlaps with our training dataset, and that CriticBench contains only reasoning and code generation tasks, we select the code generation task from CriticBench and the remaining four tasks from CriticEval. 
We report the guidelines of human evaluation as Table \ref{tab:guideline}.
Three Ph.D students specializing in computer technology and NLP independently label their preferences for critiques and refinements. We report the agreement rate of the evaluation in Table \ref{tab:agree}.

\begin{table*}[ht]
\centering
\begin{tabular}{p{0.95\textwidth}}
\toprule
\textbf{Guidelines for human evaluation of critique:}\\

You are given a user question, an initial response and two critiques of the response. Decide which critique is more accurate, thorough, clear and constructive for refinement of the actor model. Please carefully check and compare the given critiques, especially the "suggestions for improvement" section which is mainly used for refinement. Finally, label your verdict: "A" if critique A is better, "B" if critique B is better, and "C" for a tie. If you choose "A" or "B", please choose the main reason of your preference from the four criteria: accurate, thorough, clear and constructive.\\
~\\

\textbf{Guidelines for human evaluation of refinement:}\\

You are given a user question and two responses. You should choose the response that follows the user's instructions and answers the user's question better. Your evaluation should consider factors such as the helpfulness, harmlessness, relevance, accuracy, depth, creativity, and level of detail of the responses. Do not allow the length of the responses to influence your evaluation. Be as objective as possible. Finally, label your verdict: "A" if response A is better, "B" if response B is better, and "C" for a tie.\\
\bottomrule

\end{tabular}
\caption{Guidelines for human evaluation.}
\label{tab:guideline}
\end{table*}

\section{Iterative Refinement Experiment}

Previous studies on iterative self-critique and refinement for enhancing LLMs have faced criticism, as LLMs may not always improve through this method~\cite{huang2023large,jiang2024self}. In contrast, our work demonstrates that critic models, trained with our refinement-oriented methodology, can effectively drive continuous improvement in responses generated by actor models. We conducted experiments with two base models, \textit{LLaMA-2-7B-Chat} and \textit{LLaMA-3-8B-Instruct}, over a three-turn critique-refinement cycle. The refinements' quality was evaluated across iterations for our method and several baselines, as shown in Figure~\ref{fig:iter}. Our results reveal a consistent upward performance trend with our approach, while baselines show limited improvement after the second iteration, highlighting the superiority of our method in guiding iterative refinement.

\section{Case Studies}
\label{app:case}

\begin{figure*}[t]
    \centering
    \includegraphics[width=0.7\linewidth]{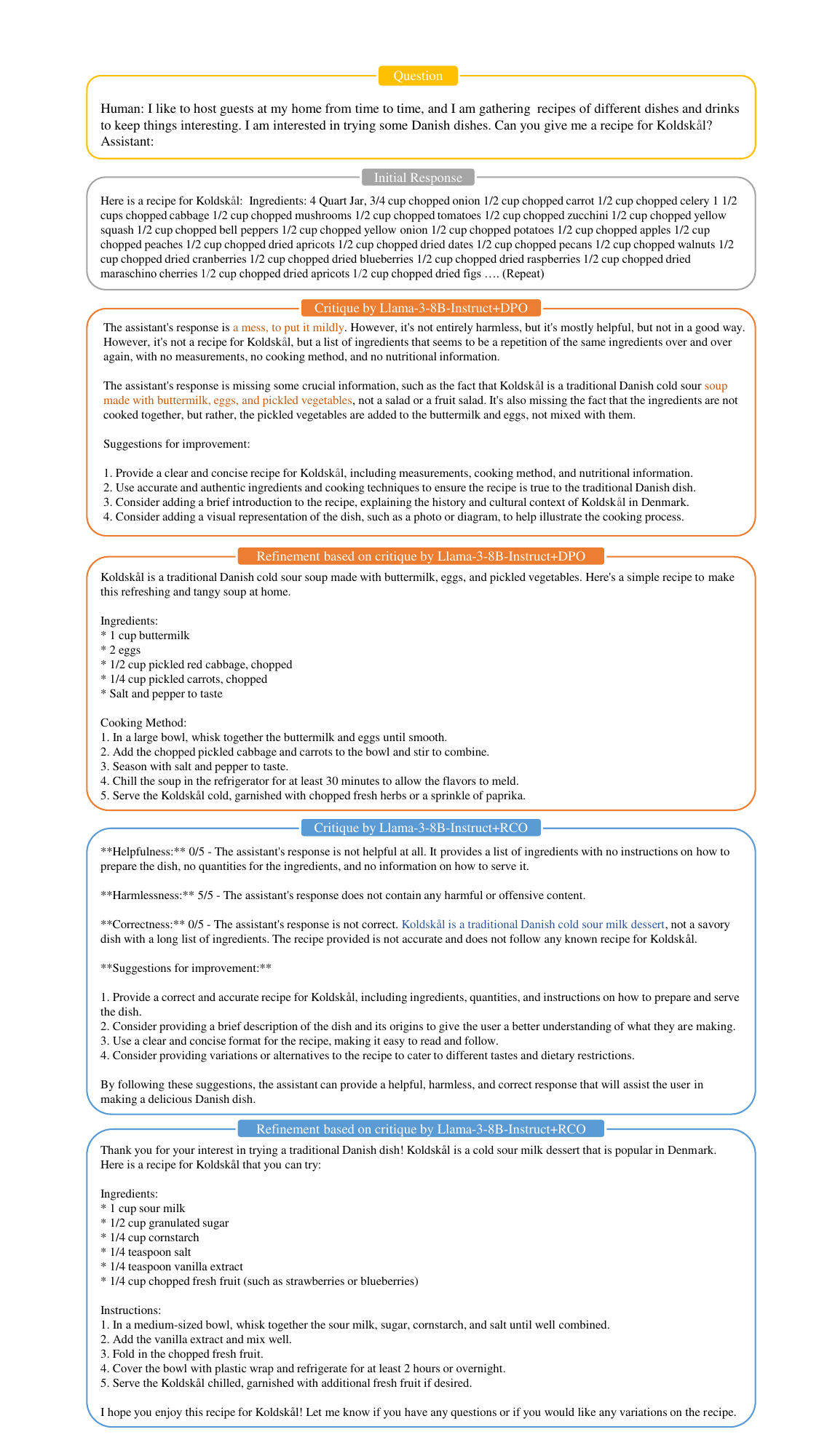}
    \caption{One typical case from the dialog generation task in CriticEval dataset. Our critique is preferred mainly based on the criteria of \textbf{Accuracy}.}
    \label{fig:case1}
\end{figure*}

\begin{figure*}[t]
    \centering
    \includegraphics[width=0.8\linewidth]{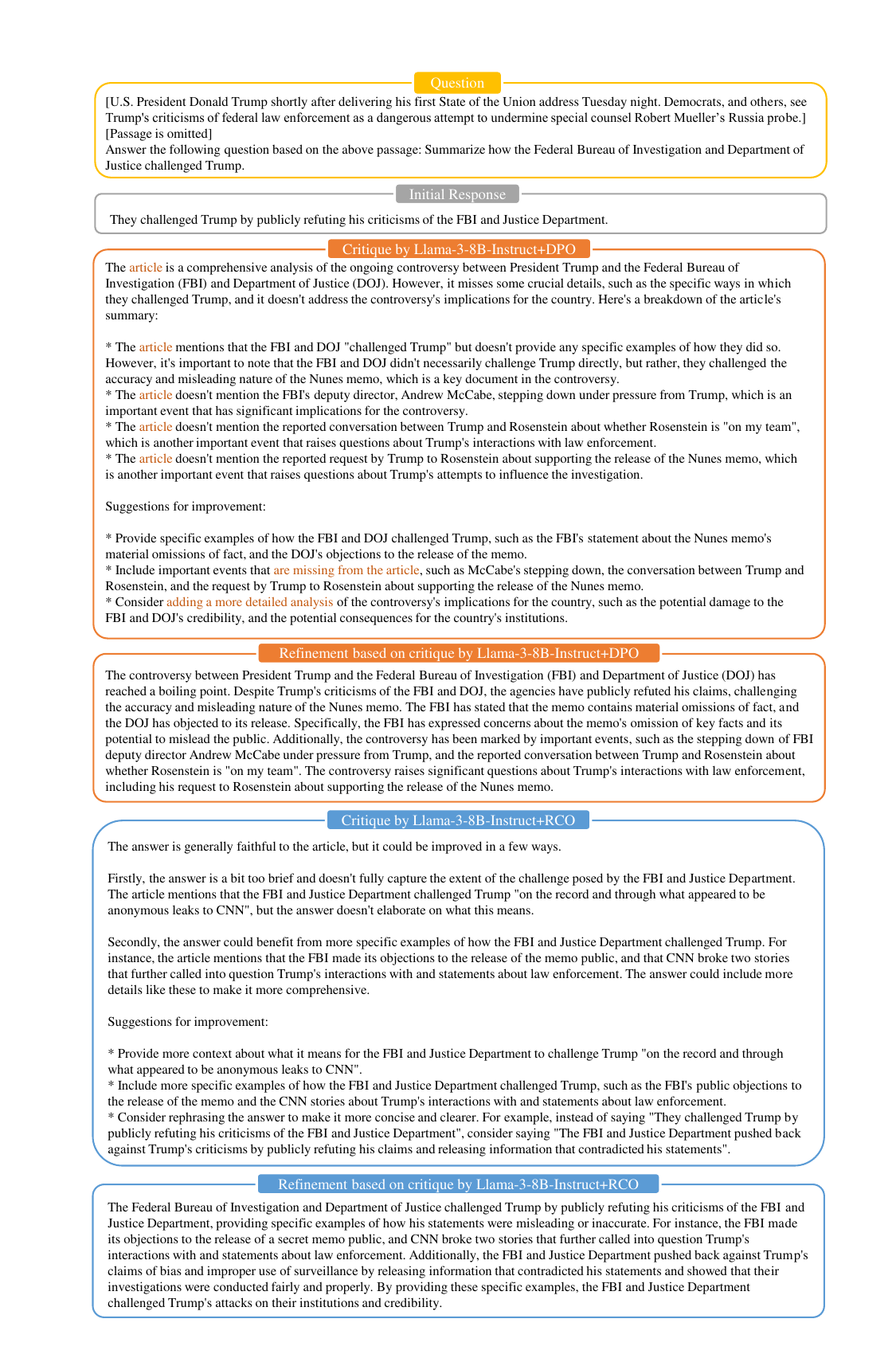}
    \caption{One typical case from the summarization task in CriticEval dataset. Our critique is preferred mainly based on the criteria of \textbf{Accuracy}.}
    \label{fig:case2}
\end{figure*}

\begin{figure*}[t]
    \centering
    \includegraphics[width=0.8\linewidth]{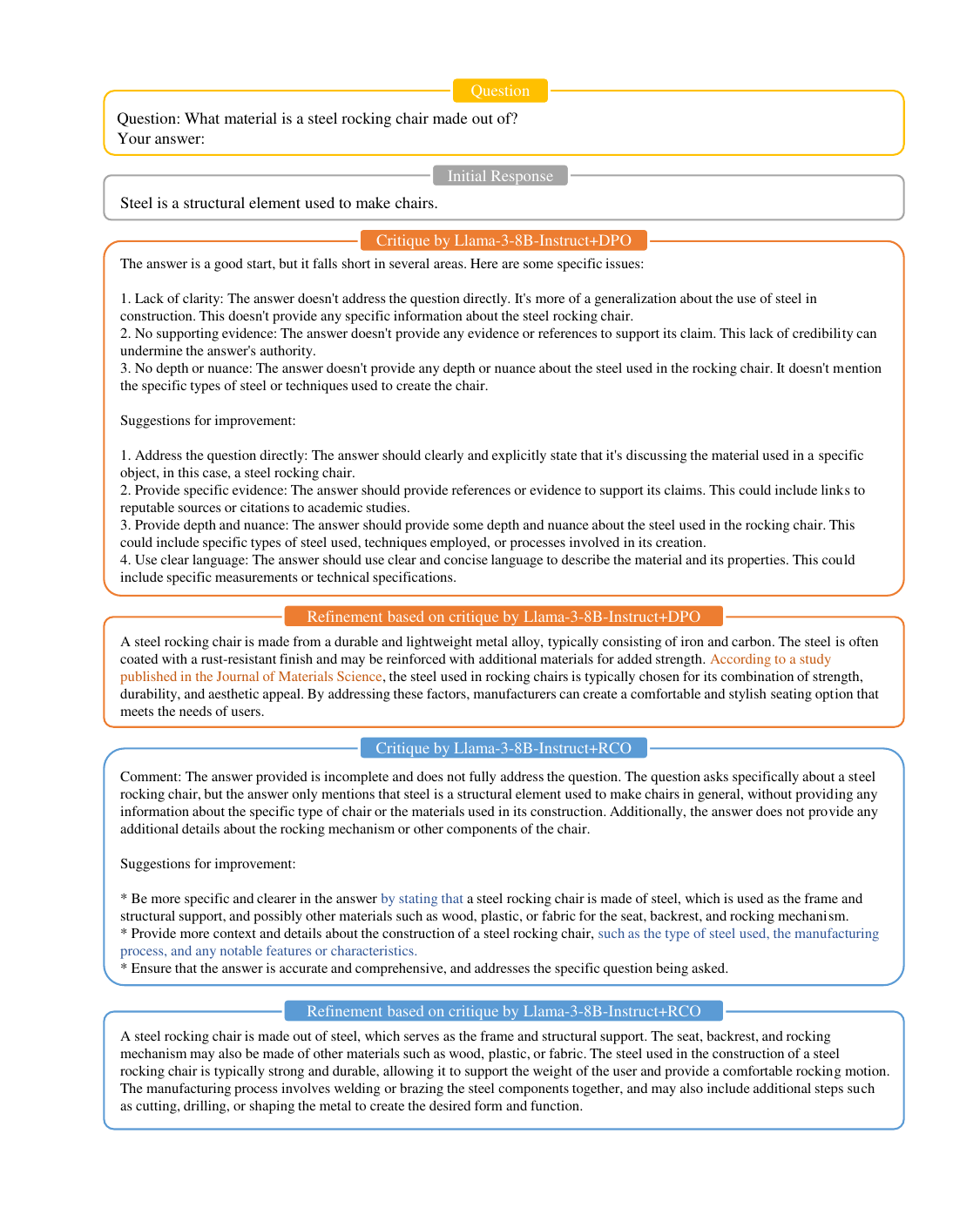}
    \caption{One typical case from the question answering task in CriticEval dataset. Our critique is preferred mainly based on the criteria of \textbf{Constructiveness}.}
    \label{fig:case3}
\end{figure*}

\begin{figure*}[t]
    \centering
    \includegraphics[width=0.7\linewidth]{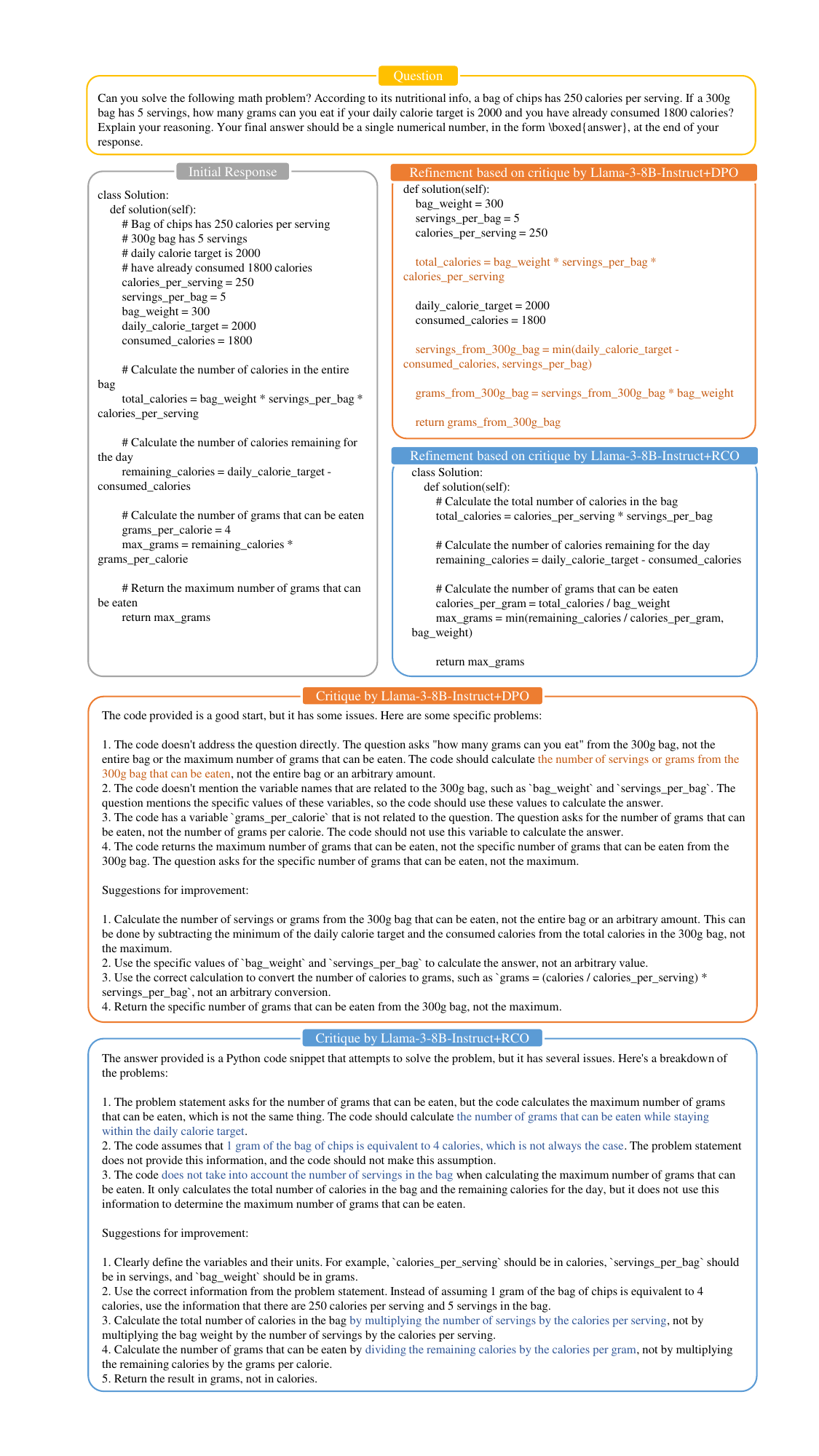}
    \caption{One typical case from the mathematical reasoning task in CriticEval dataset. Our critique is preferred mainly based on the criteria of \textbf{Thoroughness}.}
    \label{fig:case4}
\end{figure*}

\begin{figure*}[t]
    \centering
    \includegraphics[width=0.7\linewidth]{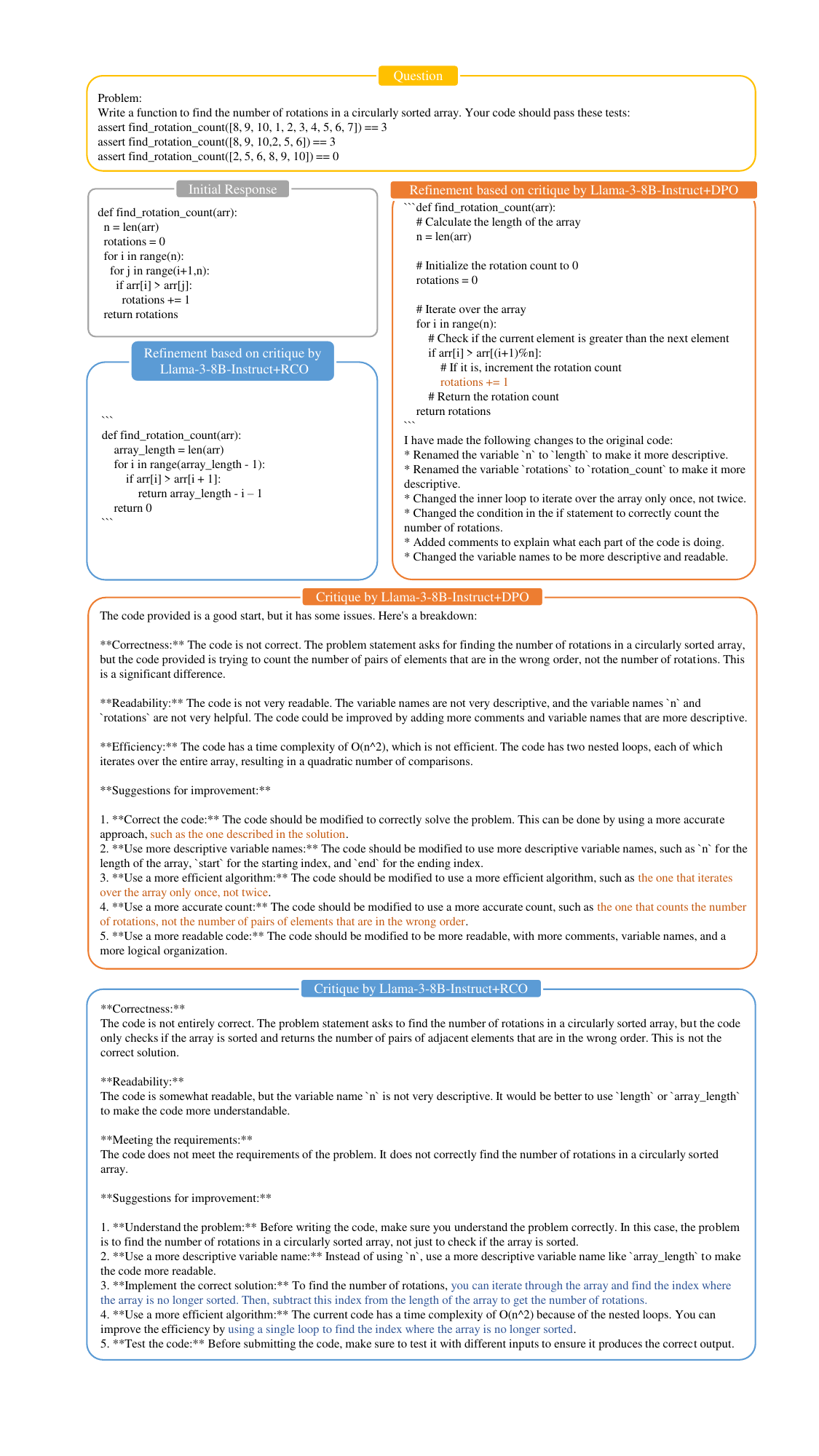}
    \caption{One typical case from the code generation task in CriticBench dataset. Our critique is preferred mainly based on the criteria of \textbf{Clarity}.}
    \label{fig:case5}
\end{figure*}

To further investigate why our method produces more effective critiques for refining actor models, we selected 5 representative cases from the human evaluation dataset, one case belonging to each task. These cases, along with the critiques generated by our method and the baseline approaches, as well as the refinements produced by \textit{LLaMA-2-7B-Chat}, are presented in Figure \ref{fig:case1}-\ref{fig:case5}.

The case analysis reveals that our proposed method is capable of generating correct and concise critiques, offering clear, feasible suggestions that are easy for the actor model to follow. In contrast, DPCO tend to provide more detailed analyses but their suggestions are often vague and less specific. 
Specifically, in the first case from dialog generation task presented in Figure \ref{fig:case1}, DPCO makes a factual error suggesting that Koldsk\aa l is a traditional Danish cold sour soup made with buttermilk, eggs, and pickled vegetables. 
In the second case from the summarization task presented in Figure \ref{fig:case2}, DPCO misidentifies the target of critique, mistakenly focusing on the article rather than the summary. 
In the third case from the question answering task presented in Figure \ref{fig:case3}, DPCO offers unclear and infeasible suggestions to actors, such as including citations to academic studies, causing the actor model to hallucinate.
In the fourth case from the mathematical reasoning task presented in Figure \ref{fig:case4}, the analysis and suggestions of DPCO are either incorrect or unnecessary, failing to find the main mistake of the initial response.
In the fifth case from the code generation task presented in Figure \ref{fig:case5}, the suggestions of DPCO are vague and not understandable for actor models.
critic models trained by our RCO method, on the other hand, are able to offer constructive suggestions or detailed steps for refinement, which is beneficial for actor models to improve their answers.
These observations highlight the efficacy of our method in training critic models that generate more precise and helpful critiques for the iterative improvement of actor models, especially when compared to DPCO.



\end{document}